\documentclass[letterpaper, 10 pt, conference]{ieeeconf}  %
\pdfoutput=1
\IEEEoverridecommandlockouts                              %
\usepackage{graphicx} %
\usepackage{times} %
\usepackage{amsmath} %
\usepackage{amssymb}  %
\usepackage{bm}
\usepackage{color}
\usepackage[belowskip=0pt]{caption}
\usepackage[hidelinks]{hyperref}
\usepackage[%
  vlined,%
  boxed%
]{algorithm2e}
\usepackage{tikz}
\usepackage{subcaption}

\hypersetup{pdfauthor={Marcus Ebner von Eschenbach, Binyamin Manela, Jan Peters, Armin Biess},pdftitle={Metric-Based Imitation Learning Between Two Dissimilar Anthropomorphic Robotic Arms}}

\makeatletter
\renewcommand{\@algocf@capt@plain}{above}
\renewcommand{\algocf@caption@plain}{\box\algocf@capbox\vskip\AlCapSkip}%
\makeatother
\SetKwProg{Fn}{Function}{}{end}\SetKwFunction{FRecurs}{FnRecursive}%

\DeclareMathOperator*{\argmin}{arg\,min} 

\title{\LARGE \bf
Metric-Based Imitation Learning Between Two Dissimilar Anthropomorphic Robotic Arms
}

\author{Marcus Ebner von Eschenbach$^{1,2}$, Binyamin Manela$^{1}$, Jan Peters$^{2}$,   Armin Biess$^{1}$%
\thanks{*This work was supported in part by the Helmsley Charitable Trust through
the Agricultural, Biological and Cognitive Robotics Initiative and the Israel Science Foundation (grant no. 1627/17).}%
\thanks{$^{1}$Department of Industrial Engineering and Management
Ben-Gurion University of the Negev, Be'er Sheva 84105, Israel
        {\tt\small abiess@bgu.ac.il}}%
\thanks{$^{2}$ Intelligent Autonomous Systems Group
Technical University Darmstadt, 64289 Darmstadt, Germany
Germany
        {\tt\small mail@jan-peters.net}}%
}

\begin{document}

\newcommand{\bea}{\begin{eqnarray}}
\newcommand{\eea}{\end{eqnarray}}
\newcommand{\nn}{\nonumber}
\newcommand{\pa}{\partial}

\tikzset{
	cframe/.pic={
		\draw [->, thick, red] (0, 0) -- (3, 0);
		\draw [->, thick, green] (0, 0) -- (0, 3);
	}
}

\maketitle
\thispagestyle{empty}
\pagestyle{empty}

\begin{abstract}
The development of autonomous robotic systems that can learn from human demonstrations to imitate a desired behavior - rather than being manually programmed - has huge technological potential. One major challenge in imitation learning is the correspondence problem: how to establish corresponding states and actions between the expert and learner, when the embodiments of the agents are different (morphology, dynamics, degrees of freedom, etc.). Many existing approaches in imitation learning circumvent the correspondence problem, for example, kinesthetic teaching or teleoperation, which are performed on the robot. In this work we explicitly address the correspondence problem by introducing a distance measure between dissimilar embodiments. This measure is then used as a loss function for static pose imitation and as a feedback signal within a model-free deep reinforcement learning framework for dynamic movement imitation between two anthropomorphic robotic arms in simulation. We find that the measure is well suited for describing the similarity between embodiments and for learning imitation policies by distance minimization.
\end{abstract}

\section{INTRODUCTION}
\begin{figure}[]
	\centering
	\vspace{0.2cm}
	\begin{subfigure}[b]{0.49\textwidth}
		\begin{tikzpicture}
		\node[anchor=south west,inner sep=0] at (0,0) {\includegraphics[width=\textwidth, trim={5cm, 6.5cm, 3cm, 8.5cm}, clip=true]{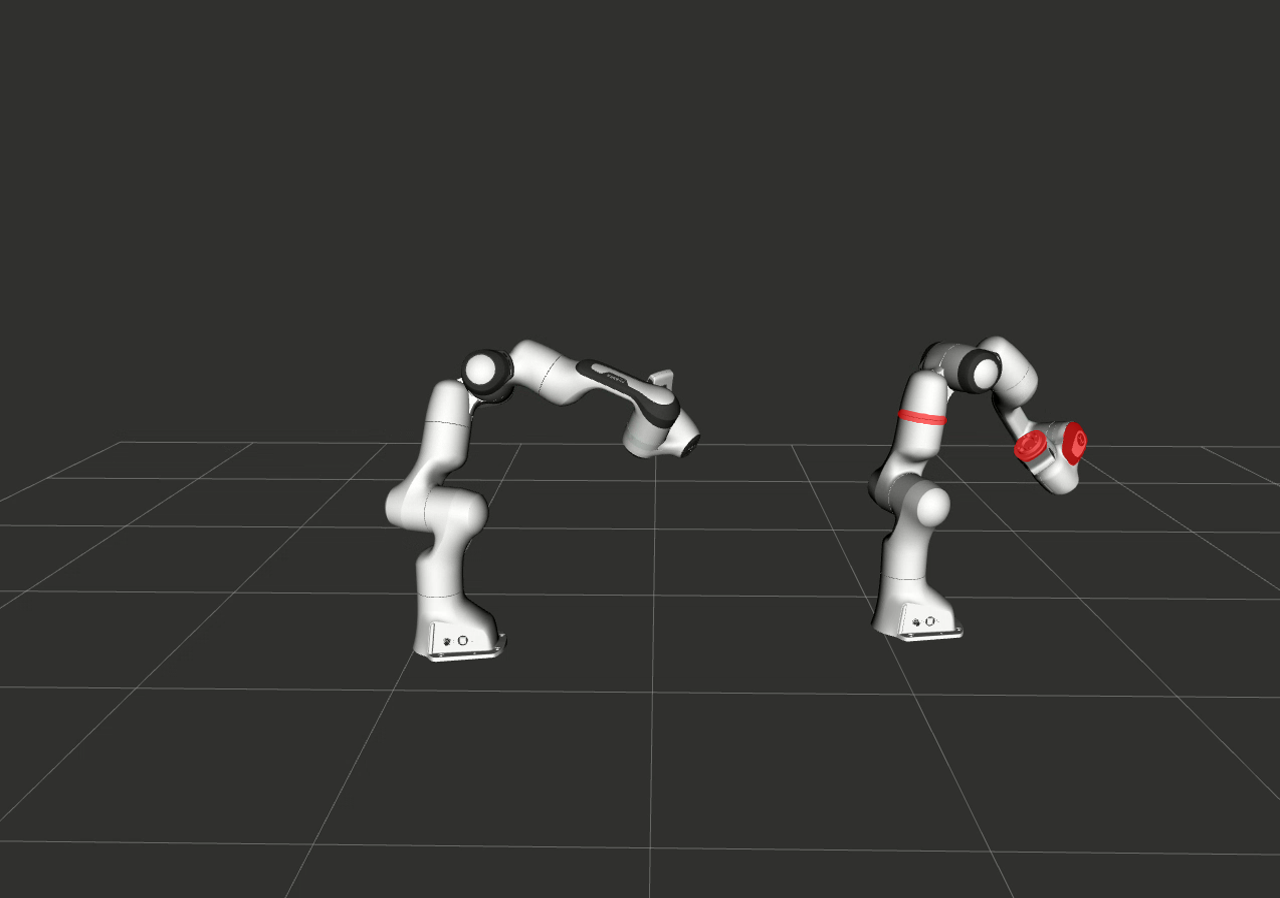}};
		\end{tikzpicture}
		\caption{}\label{fig:sgd_poses_fourjoint:a}
	\end{subfigure}
	\begin{subfigure}[b]{0.49\textwidth}
		\includegraphics[width=\textwidth, trim={6cm, 6.5cm, 2cm, 8.5cm}, clip=true]{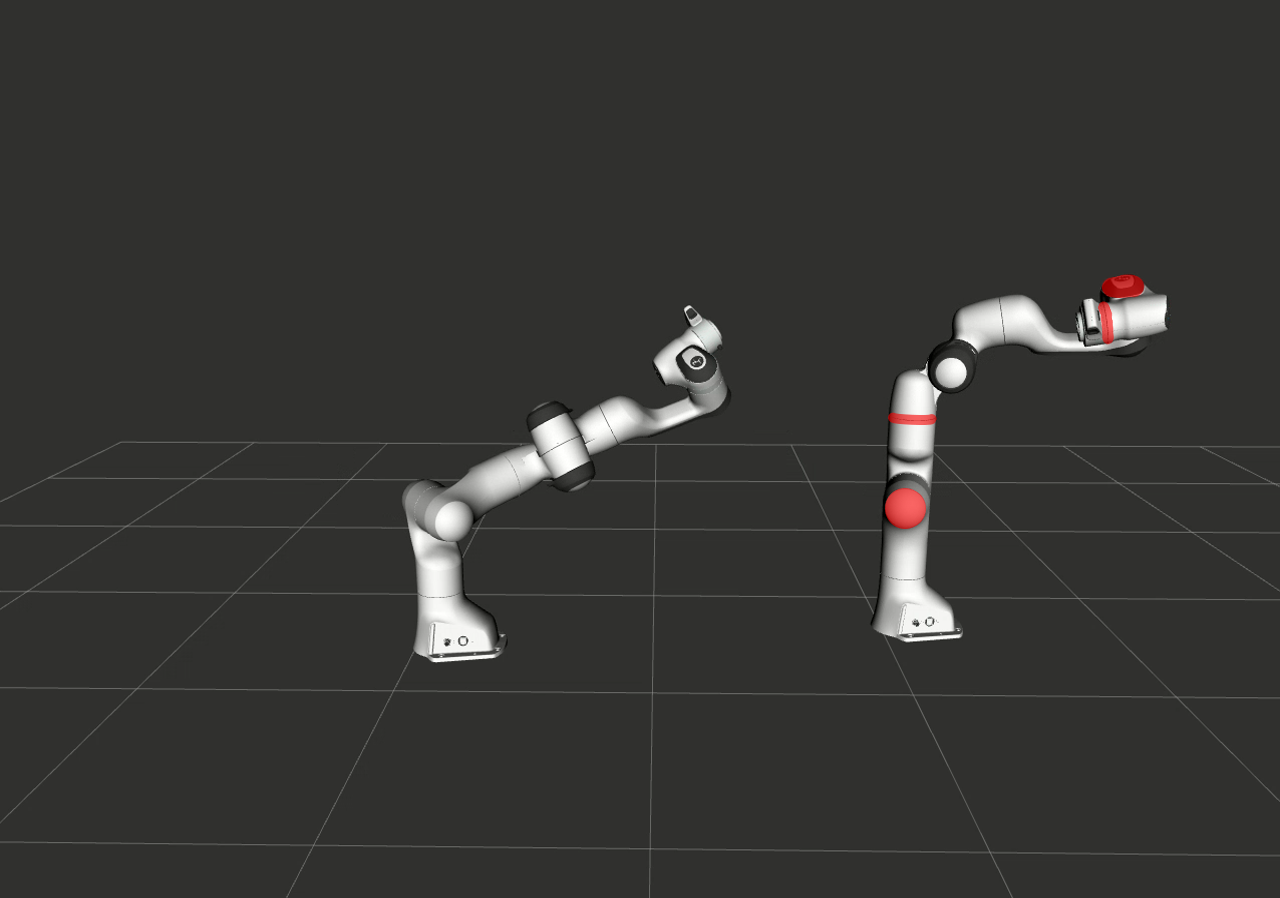}
		\caption{}\label{fig:sgd_poses_fourjoint:b}
	\end{subfigure}
	\caption[Examples of static pose imitation using a NN (expert: 7-DOF-Panda, learner: 4-DOF-Panda)]{Static pose imitation between dissimilar antrophomorphic robotic arms using a 7-DOF-Panda expert. (a) 4-DOF-Panda learner; (b) 3-DOF-Panda learner. Dissimilar robots are generated by locking DOFs (expert is shown on the left, the learner on the right, locked joints in red).}
	\label{fig:sgd_poses_fourjoint}
	\vspace{-0.2cm}
\end{figure}
Approaches to imitation learning in robotics have delivered huge success ranging from helicopter acrobatics~\cite{abbeel2010autonomous}, high-speed arm skills~\cite{kober2010imitation}, haptic control \cite{kormushev2011imitation,boularias2011relative}, gestures~\cite{calinon2010learning}, 
manipulation \cite{asfour2008imitation,lopes2007affordance,ratliff2007imitation} to legged locomotion \cite{chalodhorn2010learning,ratliff2007imitation}.
The machine learning algorithms that make imitation learning possible are well studied and have recently been summarized~\cite{osa2018algorithmic}.  Surprisingly, despite all of these impressive successes in the acquisition of new robot motor skills, fundamental research questions in imitation learning of central importance have remained open for decades.
Among such core questions is the \textit{correspondence problem}: how can one agent (the learner or imitator) produce a similar behavior - in some aspect - to behavior it perceives in another agent (the expert or demonstrator), given that the two agents obey different kinematics and dynamics (body morphology, degrees of freedom (DOFs), constraints, joints and actuators, torque limits), i.e.,  occupy different state spaces~\cite{billard2016learning}? 

Existing algorithmic approaches towards imitation learning can be divided into two groups: behavioral cloning (BC)  and inverse reinforcement learning (IRL) or inverse optimal control (IOC), both can be further subdivided into model-based and model-free approaches depending on whether the system dynamics is available or not \cite{osa2018algorithmic}.  BC and IRL make different  assumptions about the correspondence of learner and expert.  In BC, a mapping from states to actions 
is generated from the demonstrations using supervised learning methods. This mapping can then be used by the learner to reproduce similar behavior, provided that the embodiments of expert and learner are alike, otherwise the method will fail due to lack of correspondence. 

Successful implementations of model-based BC algorithms have been obtained for a hitting-a-ball task with an underactuated robot~\cite{englert2013probabilistic}, playing video games~\cite{ross2011reduction} and controlling a UAV through a real forest~\cite{ross2013learning}. Model-free BC algorithms have been implemented for autonomous  RC helicopter flight using experts demonstrations~\cite{abbeel2010autonomous} and for learning tasks, such as tennis swings ~\cite{ijspeert2002movement},  ball-paddling~\cite{kober2010imitation} and human-robot collaborative motions in tool-handover tasks~\cite{maeda2017phase}, autonomous knot-tying using a surgical robot~\cite{osa2014trajectory} and character animation~\cite{peng2018deepmimic}.

In an IRL framework, the learner infers a reward function for a given task from expert demonstrations of the task. The underlying assumption is that the reward function is a parsimonious and portable representation of the task, which can be transferred and generalized to agents with different embodiments. Thus, IRL implicitly resolves the correspondence problem but has the disadvantage of being computationally expensive and requires reinforcement learning in an inner loop.  IRL has been implemented mostly in a model-based approach for tasks, such as learning to drive a car in a simulator \cite{abbeel2004apprenticeship} and path planning \cite{ratliff2006margin,silver2010navigation,ziebart2008entropy}. A few model-free IRL algorithms  have been proposed and used to learn policies of robot motion \cite{boularias2011relative,finn2016guided}, {\cite{ho2016model}}.

The correspondence problem results in the  following question: what action sequence in the learner is required  to produce  behavior that is similar to the expert, given that learner and expert have different embodiments and given that a measure of similarity $d$ is defined.  If we denote the state-action pairs of the expert and learner as  $(\bm{s}_t,\bm{a}_t) $ and  $(\bm{\hat{s}}_t,\bm{\hat{a}}_t)$, respectively, with $t=1,2,\dots T$, then we can formulate the correspondence problem in its simplest form as follows: for a given set of demonstrations  ${\cal{D}} = \{\bm{s}_1,\bm{a}_1,\dots, \bm{a}_{T-1},\bm{s}_T\}_{i=1}^N$  find actions $\bm{\hat{a}}_t, t=1,2,\dots T$, so that $\bm{\hat{s}}_t$ is similar to $\bm{s}_t$ for all $t=1,2,\dots T$, where similarity (negative loss) is defined as $\sum_{t=1}^T d(\bm{s}_t,\bm{\hat{s}}_t)~\rightarrow~\mbox{min}$.   Note that the states depend on the actions via the system dynamics, thus, it is $\bm{{s}}_{t+1} = \bm{f}(\bm{{s}}_t,\bm{{a}}_t) $ and $\hat{\bm{s}}_{t+1} = \hat{\bm{f}}(\hat{\bm{s}}_t,\hat{\bm{a}}_t)$, where $\bm{f}$ and $\hat{\bm{f}}$ describe the system dynamics of expert and learner, respectively.  Several levels of complexity can be added to this formulation. (1) Generally, the dynamics for real systems are stochastic, thus, $\bm{{s}}_{t+1} = \bm{f}(\bm{{s}}_t,\bm{{a}}_t) + \bm{\epsilon}_t$ and $\hat{\bm{s}}_{t+1} = \hat{\bm{f}}(\hat{\bm{s}}_t,\hat{\bm{a}}_t)+ \hat{\bm{\epsilon}}_t$, where $\bm{\epsilon}_t$ and $\hat{\bm{\epsilon}}_t$ is noise. In this case the distance can be measured between probability distributions of state vectors,  $\sum_{t=1}^T d(p(\bm{s}_t),\hat{p}(\bm{\hat{s}}_t))\rightarrow \mbox{min}$, where $p(\bm{s}_t)$ and $\hat{p}(\hat{\bm{s}}_t)$ denote the probability distribution for expert and learner, respectively  (2) The system dynamics of expert and/or learner are often not known, leading to model-free vs. model-based approaches.  (3) The demonstrations are often given  in the form ${\cal{D}} = \{\bm{s}_1,\dots, \bm{s}_T\}$, i.e., the actions of the expert are not available. (4) The states of the expert may be only partially observable, for instance, if the environment is observed  by cameras. The states must then be inferred from observations $\bm{o}_t$ only. 

In this work we study imitation tasks between two dissimilar anthropomorphic robot arms, which are generated by locking degrees of freedom (DOFs) in the learner. Throughout this work we assume that the dynamics of the learner are not available to the learning agent, thus, we are in a model-free setting. %
We first introduce our definition of an embodiment state and provide a distance measure to assess the similarity between embodiments. This distance measure is then used to imitate  static poses using neural networks (Fig.\ref{fig:sgd_poses_fourjoint}) and as a feedback signal for movement imitation using reinforcement learning.
\section{RELATED WORK}
Metric approaches to the correspondence problem in imitation have been developed in a series of studies~\cite{nehaniv2001likeme,alissandrakis2002alice,alissandrakis2002do,alissandrakis2005corresponding,
alissandrakis2007correspondence,nehaniv2007book}. In these studies, the correspondence problem was formulated in state-action space with separate metrics for states and actions. Simple global metrics based on the  Hamming norm, Euclidean distance and infinity norm ($L_p$-norms) were used to measure the similarity between expert and learner.  
 Another approach to the correspondence problem is to explicitly learn the forward dynamics of the learner. The actions of the learner are then adapted to the given demonstrations by using the learned forward dynamics. Within this framework, Englert et al.~\cite{englert2013probabilistic} have used the Kullback-Leibler divergence as similarity measure to compare trajectory distributions from an expert and a robot learner.   Similarly, Grimes et al.~\cite{grimes2006dynamic,grimes2009learning} have used 
Gaussian Mixture Models  to learn a forward model and infer optimal actions of the learner. The model has been used to transfer human motions to a humanoid robot.

\section{METHODS}

\subsection{Definition of an embodiment}
In this study an embodiment consists of a chain of links, which are represented by frames that are attached to each link. Frames are commonly used in robotics to describe the orientation and translation (pose) of rigid bodies with respect to a laboratory frame. Frames are elements of the special Euclidean group $SE(3)$, which is a non-Euclidean manifold and group (Lie-group).  Frames can be represented as homogeneous matrices defined as
\bea
\bm{T} = \begin{bmatrix} \bm{R} & \bm{p}\\ \bm{0}^T & 1 \end{bmatrix}\,, \label{0}
\eea
where $\bm{R} \in SO(3)$ is a rotation matrix ($\bm{R}\bm{R}^T =\bm{R}^T \bm{R} =\bm{I}, \det \bm{R}=1$) and $\bm{p} \in \mathbb{R}^{3}$ is a column vector, describing the orientation and translation of a frame, respectively, with respect to a reference frame. For simplicity we write $\bm{T}=[\bm{R},\bm{p}]$. The inverse is then defined as $\bm{T}^{-1}=[\bm{R}^T,-\bm{R}^T\bm{p}]$.  The configuration space of an embodiment consisting of $n$ links with attached frames can be described by an element of the direct product space  $SE(3)^n = SE(3)\times SE(3) \times \dots \times SE(3)$ ($n$ copies).  The velocity of a frame is described by a twist, which encodes the rotational  and  translational velocity of the rigid body.  A twist is an element of the Lie-algebra $se(3)$, which defines a vector space. The velocity of the embodiment consisting of $n$ frames is described by an element of the direct product space  $se(3)^n = se(3)\times se(3) \times \dots \times se(3)$ ($n$ copies). A twist can be represented by $4\times4$ matrices of the form
\bea
{\cal{\bm{V}}}^b = \bm{T}^{-1}\dot{\bm{T}}= \begin{bmatrix} [{\bm{\omega}}^b]& \bm{v}^b\\ \bm{0}^T & 0 \end{bmatrix}
\label{0a}
\eea
or

\bea
{\cal{\bm{V}}}^s = \dot{\bm{T}}\bm{T}^{-1} = \begin{bmatrix} [{\bm{\omega}}^s] & \bm{v}^s\\ \bm{0}^T & 0 \end{bmatrix}\;,
\label{eq:spatialtwist}
\eea
where (\ref{0a}) defines the body twist and (\ref{eq:spatialtwist}) the spatial twist. 
The notation $[\cdot]$ denotes a skew symmetric $3\times 3$ matrix composed of the components of the angular velocity $\bm{\omega} = [\omega_{1},\omega_{2},\omega_{3}]^T$, that is
\bea
[\bm{\omega}] = -[\bm{\omega}]^T = \begin{bmatrix}
0 & -\omega_{3} & \omega_{2} \\
\omega_{3} & 0 & 	-\omega_{1} \\
-\omega_{2} & \omega_{1} & 0
\end{bmatrix}\,.
\eea

 Specifically, $[\bm{\omega}^b]=\bm{R}^T\dot{\bm{R}} \in \mathbb{R}^{3\times3}$ and  $\bm{v}^b=\bm{R}^T\dot{\bm{p}} \in \mathbb{R}^3$ define the angular velocity and translational velocity of the origin with respect to the base frame, respectively, both expressed in coordinates of the body frame. A similar but less intuitive physical interpretation can be given to the spatial twist.  For simplicity we write ${\cal{\bm{V}}} = [\bm{\omega},\bm{v}]$. 
The joint of the first link is always attached to the origin of the \textit{base frame}, which serves as a reference frame in which all comparisons will be performed. Each joint rotates around one axis. The forward kinematic map is a map from joint angles ${\bm{q} = [q_1, q_2, \dots, q_n]^T}$
to frames $\bm{q} \rightarrow \bm{T}(\bm{q})$, where $n$ denotes the number of DOFs. For simplicity we first consider a planar manipulator with $n$ DOFs.  We assume that all links have cylindrical shape and constant mass density. We attach a frame to each link $i$  with its origin  at a distance of $r_i$ from joint $i-1$,   
 $i=1,\dots, n,$ and with the $x$-axis that is pointing along the link direction. The transformation from link frame $i$ to the base frame $0$ can then be described by a product of matrix exponentials 
\bea
\bm{T}_{0i}(q_1, q_2, \dots, q_{i})= e^{q_1 {{\bm{S}}}_{1}}e^{q_2 {{\bm{S}}}_{2}}\dots e^{q_i {{\bm{S}}}_{i}}\bm{M}_{i}\,,
\label{1}
\eea
where 
\bea
\bm{M}_{i} = \begin{bmatrix} 1 & 0 & 0 & r_i\\ 0 & 1& 0 & 0 \\ 0 & 0 & 1 & 0 \\ 0 & 0 & 0 & 1\end{bmatrix}\,,\;\; \label{5a} {{\bm{S}}}_{i}  = \begin{bmatrix}[{\bm{n}}_{i}]& -\bm{n}_{i}\times \bm{q}_{i} \cr 0&0 \cr \end{bmatrix},
\eea
for $i=1,\dots,n$. The homogeneous matrix $\bm{M}_{i}$ describes a constant shift 
of the frame $i$ by $r_i$ along the $x$-axis. The screw $S_{i} \in se(3)$ is a $4\times 4$ matrix and describes the rotation axis of the revolute joint $i$. Here  $\bm{n}_{i}$ denotes a unit vector in the direction of the joint axis $i$, $\bm{q}_{i}$ is a vector from the base to any point on the joint axis $i$ (both expressed in coordinates with respect to the base frame) and $\times$ denotes the vector cross product.
Common choices for $r_i$ are  $r_i =  l_i/2$, which corresponds to attaching frames  to the center of mass (COM)  of each link (our choice) and $r_i =  l_i$, which corresponds to attaching frames  to the end of each link.
 Joint angles and joint velocities ${\dot{\bm{q}} = [\dot{q}_1, \dot{q}_2, \dots, \dot{q}_n]^T}$ determine  the twists, thus
 $(\bm{q},\dot{\bm{q}}) \rightarrow {\cal{\bm{V}}}(\bm{q},\dot{\bm{q}})$. The (body) twists  follow from (\ref{0a}) and  (\ref{1})  as ($i=1,\dots,n$)
 \bea
{\cal{\bm{V}}}_{0i} = \bm{T}^{-1}_{0i}(q_1, q_2, \dots, q_{i})\cdot \dot{\bm{T}}_{0i}(q_1, q_2, \dots, q_{i})\,, \label{6}
\eea
which can be determined recursively leading to
\bea
{\cal{\bm{V}}}_{0i} = \mbox{Ad}_{[\bm{M}_i e^{\bm{S}_i q_i}]^{-1}} ({\cal{\bm{V}}}_{0i-1})  + \bm{S}_i\dot{q}_i\,,
\eea
where the adjoint map $\mbox{Ad}_{\bm{T}}({\cal{\bm{V}}})$ for frame $\bm{T}=[\bm{R},\bm{p}]$ and twist ${\cal{\bm{V}}}=[\bm{\omega},\bm{v}]$  is defined as $\mbox{Ad}_{\bm{T}}({\cal{\bm{V}}}) = [\bm{R} \bm{\omega}, \bm{p} \times \bm{R} \bm{\omega} + \bm{R} \bm{v}] \in se(3)$ and ${\cal{\bm{V}}}_{00} :=\bm{0}$ \cite{parkLie}.
Note that from the (body) twists ${\cal{\bm{V}}}_{0i}= [\bm{\omega}_{0i},\bm{v}_{0i}]$ the angular velocity and translational velocity of frame $\bm{T}_{0i} = [\bm{R}_{0i},\bm{p}_{0i}]$ with respect to the base frame can be easily obtained by
\bea
\bm{\omega}_{0i}^s &=& \bm{R}_{0i}\bm{\omega}_{0i}\,, \label{9a}\\
\dot{\bm{p}}_{0i} &=& \bm{R}_{0i}\bm{v}_{0i}\,.\label{9b}
\eea
For further details using frames and twists we refer to \cite{lynch2017modern}.
After these derivations we can define the state of an embodiment $\bm{s}$ as the state of all frames 
\bea
\bm{s} &=& (\bm{s}_1,\dots , \bm{s}_i,\dots, \bm{s}_n)\nn\\
&=& (\bm{T}_{01},{\cal{\bm{V}}}_{01},\dots \,, \bm{T}_{0i},{\cal{\bm{V}}}_{0i}, \dots \,, \bm{T}_{0n},{\cal{\bm{V}}}_{0n} )\,, \label{4}
\eea
where we surpressed the arguments and $\bm{s}_i=(\bm{T}_{0i},{\cal{\bm{V}}}_{0i}) \in \bm{{\cal{S}}}_i$ denotes the state of frame $i$. Note  that via the foward kinematic map, which is assumed  to be known in this work,  an embodiment state is fully determined by its joint angles $\bm{q}$ and joint velocities $\dot{\bm{q}}$, i.e., $\bm{s} = \bm{s}(\bm{q},\dot{\bm{q}})$. A special case of (\ref{4}) is obtained when ignoring rotational information. In this case, an embodiment can be described by the position and (translational) velocity of a set of candidate points, defined by the origin of each frame  (by setting $\bm{R}=\bm{I}$ in (\ref{0}), (\ref{0a})).  The embodiment state is then described by  
\bea
\bm{s} &=& (\bm{x}_1,\bm{v}_1, \dots \,, \bm{x}_i,\bm{v}_i,\dots \,, \bm{x}_n,\bm{v}_n )\,, \label{5}
\eea
where $\bm{s}_i=(\bm{x}_i,\bm{v}_i)$ denotes the state vector of candidate point $i$. The definition of an embodiment in terms of frames/twists and candidate points is generic and can be applied to any robot. A disadvantage of using frames is that they are not elements of a vector space, but define a non-Euclidean manifold.

\subsection{Similarity between embodiments}
\begin{figure}[]
	\centering
	\begin{subfigure}[b]{0.357\linewidth}
		\includegraphics[width=\textwidth]{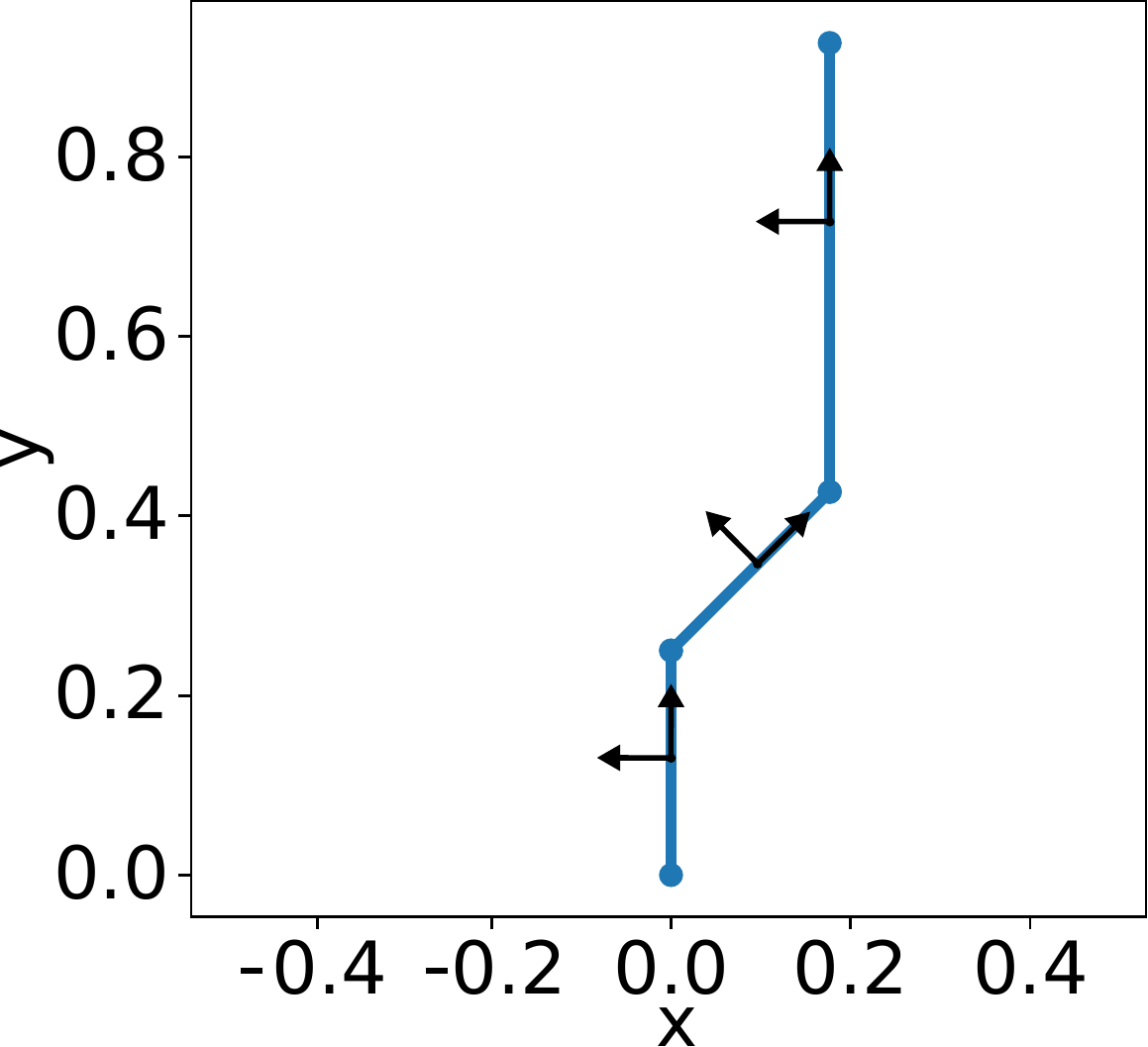}
		\caption{} \label{fig:whenembodimentssimilar:a}
	\end{subfigure}
	\begin{subfigure}[b]{0.3\linewidth}
		\includegraphics[width=\textwidth]{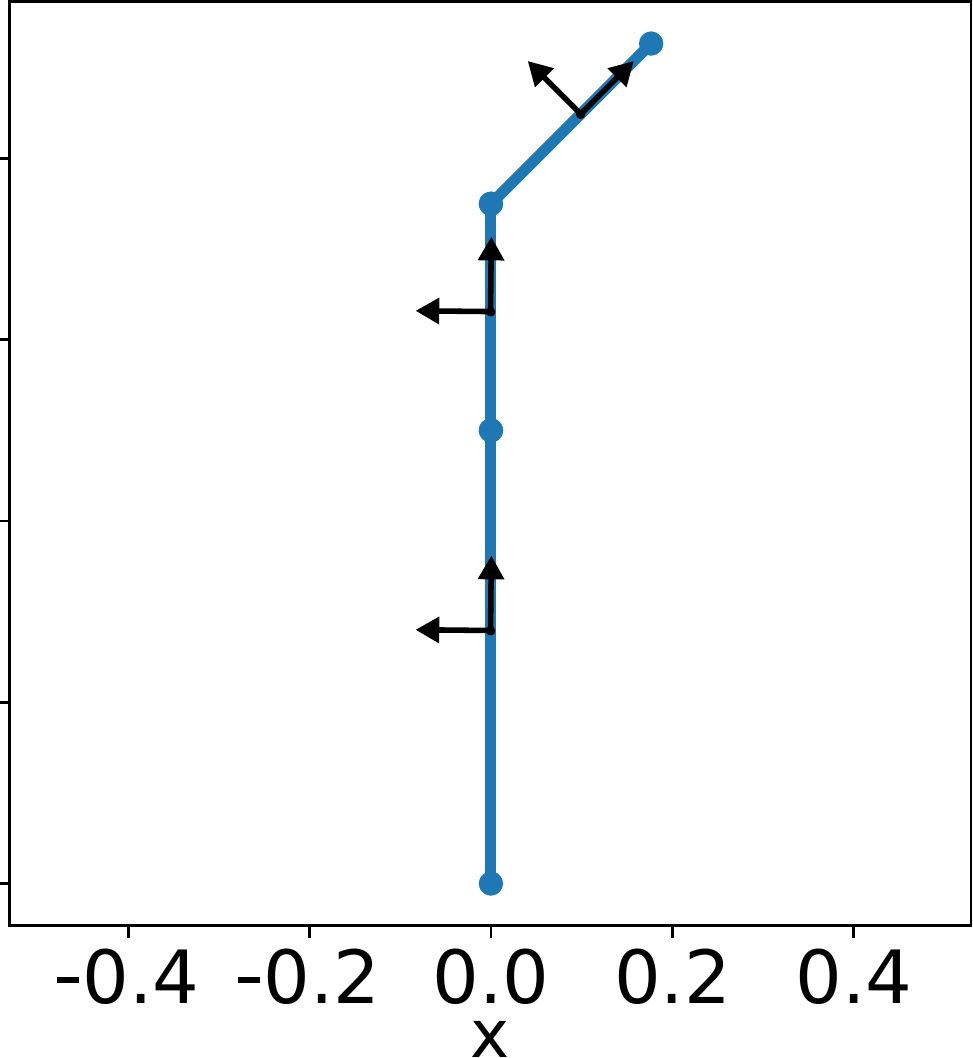}
		\caption{} \label{fig:whenembodimentssimilar:b}
	\end{subfigure}
	\begin{subfigure}[b]{0.3\linewidth}
		\includegraphics[width=\textwidth]{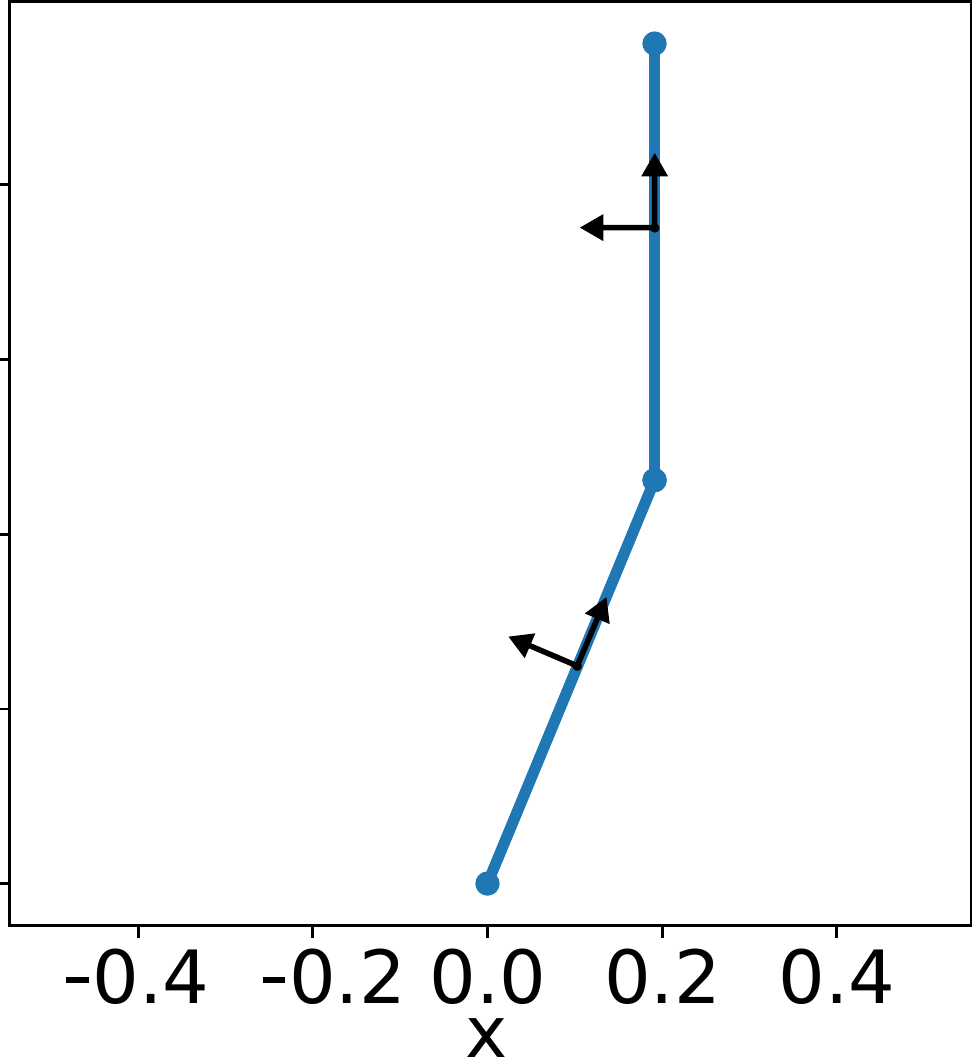}
		\caption{} \label{fig:whenembodimentssimilar:c}
	\end{subfigure}
	\caption{Pose imitation task between planar manipulators. (a) Demonstrator pose; (b, c) Learner poses. Learner (b) generates  perfect imitation if similarity is defined by candidate points, e.g., end-effector position. However, for a similarity measure based on frames, the two embodiments do not resemble each other. Learner (c) -- consisting only of two links  -- provides better imitation than learner (b) if similarity is measured between frames attached to each link.}
	\label{fig:whenembodimentssimilar}
	\vspace{-0.2cm}
\end{figure}
Similarity of embodiments can be assessed by defining a distance measure $d$ between frames/twists or candidate points, that is
\begin{equation*}
d(\bm{s}_i, \hat{\bm{s}}_j): \bm{{\cal{S}}}_i  \times   \hat{\bm{{\cal{S}}}}_j \mapsto \mathbb{R}^+_0\,,
\end{equation*}
where $\bm{s}_i \in \bm{{\cal{S}}}_i $ and $\hat{\bm{s}}_j \in 
\hat{\bm{{\cal{S}}}}_j$ result from different state spaces. We first consider the distance
\bea
d(\bm{T}_i,\hat{\bm{T}}_j) &=& \alpha_{tr}d_{tr}(\bm{T}_i,\hat{\bm{T}}_j) + \alpha_{rot}d_{rot}(\bm{T}_i,\hat{\bm{T}}_j) \label{8}
\eea
between two frames, $\bm{T}_i=[\bm{R}_i,\bm{p}_i]$ and $\hat{\bm{T}}_j=[\hat{\bm{R}}_j,\hat{\bm{p}}_j]$. The distance consists of a translational and rotational part, which are weighted with factors $\alpha_{tr}$ and $\alpha_{rot}$. The weights can be either constants or functions of other variables. For the translational part we set the Euclidean distance between the two frame origins
\bea
d_{tr}(\bm{T}_i,\hat{\bm{T}}_j)  = \lVert{\bm{p}_i - \hat{\bm{p}}}_j\lVert \,.
\label{14}\eea
There are various ways to define the rotational distance between frames. We choose to take the angle between the unit vectors pointing along the $x$-axes of the frames, i.e., into the directions of the links. Thus, we define
\bea
\beta = \arccos(\bm{e}_x^i\cdot\hat{\bm{e}}_x^j)
\eea
leading to values in the interval $[0, \pi]$.
This definition results in numerical problems when performing gradient descent because the derivative of the $\arccos$-function is
\bea
\frac{d}{dx}\arccos x = - \frac{1}{\sqrt{1-x^2}}\,.
\eea
To avoid singularities, a modified rotational distance can be defined by shifting the negated $\cos \beta$ into the interval~$[0, \pi]$ as
\bea
d_{rot}(\bm{T}_i,\hat{\bm{T}}_j) = \frac{\pi}{2}(1 - \cos\beta) = \frac{\pi}{2}(1 - \bm{e}_x^i\cdot\hat{\bm{e}}_x^j)\,. \label{17}
\eea
Note that the direction of the $x$-axis of frame $\bm{T}_{0i} =  [\bm{R}_{0i},\bm{p}_{0i}]$ with respect to the laboratory frame can be easily extracted as the first column of the rotation matrix $\bm{R}_{0i}$.  The distance measure introduced in (\ref{8}), (\ref{14}) and (\ref{17}) includes only the static pose of the embodiment and frames might be considered similar, even though they move into different directions. To include also motion information into the distance measure, the twists of the frames need to be taken into consideration. For dynamic motion imitation  we therefore augment the distance measure between two states $\bm{s}_i = (\bm{T}_i, {\cal{\bm{V}}}_i) \in \bm{{\cal{S}}}_i$ and $\hat{\bm{s}}_j= (\hat{\bm{T}}_j, {\hat{\cal{\bm{V}}}}_j) \in \hat{\bm{{\cal{S}}}}_j$ by including the translational and angular velocity (\ref{9a}) and (\ref{9b}), that is
\bea
d(\bm{s}_i,\hat{\bm{s}}_j) = \alpha_{tr}d_{tr}+ \alpha_{rot}d_{rot}+ \alpha_{v}d_{v} + \alpha_{\omega}d_{\omega} \,,
\eea
with 
\bea
d_{v} = \lVert \dot{\bm{p}}_i - \dot{\hat{\bm{p}}}_j \lVert \;\;\;\; \mbox{and} \;\;\;\;
d_{\omega} = \lVert \bm{\omega}^s_i -\hat{\bm{\omega}}^s_j \lVert\,.
\eea
Note that the distance measure between two states $\bm{s}_i$ and $\hat{\bm{s}}_j$, can be extended over all state spaces by defining the sum of all mutual distances
\bea
d(\bm{s},\hat{\bm{s}}) = %
 \sum_{i=1}^{n}\sum_{j=1}^{\hat{n}}  d(\bm{s}_i,\hat{\bm{s}}_j)\,.\label{21}
\eea
In the next section a weighted version of (\ref{21})  will be introduced by incorporating link correspondences.

\subsection{Link Correspondences}
\label{sec:link_assignment}
To measure similarity between links of two  embodiments, we first define how  correspondence between links of different embodiments can be established. Embodiments may differ in the number and length of links, and thus, a one-to-one assignment between links is often not possible. 
To establish correspondence between links of different embodiments with possibly different overall size, we first rescale each embodiment by the sum of its link lengths $L$, resulting in a chain length of $1$. 
To establish correspondence, we assign a weight for every possible link-pair combination. Thus, for two embodiments 1 and 2 with $n$ and $\hat{n}$ links, respectively, link correspondence can be represented by a correspondence matrix $\bm{W}\in \mathbb{R}^{n \times \hat{n}}$. Irrelevant combinations result in zero  or close to zero entries and higher values indicate higher relevance.
Each row of the correspondence matrix contains the correspondence weights of one link of embodiment 1 to all other links of embodiment 2, where the highest value indicates which link is the most relevant from all links of embodiment 2. %
The elements of the correspondence matrix can either be calculated as a function of embodiment states or pre-calculated for a pair of embodiments, independent of their current state.

\textit{State-Dependent Assignment}. For state-dependent calculations of the correspondence matrix $\bm{W}(\bm{s}, \hat{\bm{s}})$, weights are calculated 
using the distance measure between frames. The closer the distance,
the higher the weight for this pair of frames. To obtain the correspondence matrix $\bm{W}(\bm{s}, \hat{\bm{s}})$,  the mutual  distance matrix $\bm{D}'(\bm{s}, \hat{\bm{s}}) = ({D}'(\bm{s}_i, \hat{\bm{s}}_j))\,, i=1\dots n, j=1\dots \hat{n},$ between all links of the two embodiments is computed using the distance measure in (\ref{8}). A correspondence matrix can be generated by replacing the smallest element of each row of $\bm{D}'(\bm{s}, \hat{\bm{s}})$ with 1 and all other elements with 0, resulting in a binary matrix $\bm{W}_{12}(\bm{s}, \hat{\bm{s}})$ that assigns exactly one link of embodiment 1 to each link of embodiment 2  with a weight 1.
The same operation can be applied for each column of $\bm{D}'(\bm{s}, \hat{\bm{s}})$, resulting in $\bm{W}_{21}$. Adding~$\bm{W}_{12}$ to~$\bm{W}_{21}$ results in a correspondence matrix~$\bm{W}$. 
A correspondence matrix that only uses the minimum for each row and each column is very selective and ignores the fact that more than one link of the other embodiment may be lying in a similar distance and should be taken into consideration. This effect can be mitigated by applying a %
\textit{softmax} function to the rows and columns of the correspondence matrix, after multiplying with a constant factor $\xi < 0$ to find soft \textit{minima} instead of maxima and to adjust the distinctness of the minimum.

\subsection{Calculating Distance Between Embodiments}
We define distance between embodiments as 
 \textit{element-wise} multiplication of the distance function with the correspondence matrix  i.e., $\bm{D}(\bm{s}, \hat{\bm{s}}) =\bm{W}(\bm{s}, \hat{\bm{s}}) \circ \bm{D}'(\bm{s}, \hat{\bm{s}})$, where~$\circ$ denotes the Hadamard product.    Only  distances between corresponding link pairs remain because non-corresponding pairs are weighted with zero or near-zero values. To obtain one single scalar number, the mean of all entries of the resulting matrix is taken. Matrix norms, such as the Frobenius norm, can also be used.
For the evaluation of the correspondence matrix and the distance matrix, suitable weights, $\alpha_{tr}$ and $\alpha_{rot}$, need to be chosen. Different settings are possible here. 
The pseudo-code for calculating the distance measure are shown in Algorithm~\ref{alg:distancemeasure}.
\begin{algorithm}
	\DontPrintSemicolon
	\Fn{distance\_measure($\bm{s}(\bm{q})$, $\hat{\bm{s}}(\hat{\bm{q}})$, $\alpha_{tr}$, $\alpha_{rot}$)}{
		Calculate $\{\bm{T}_i\}_{i=1}^{n}, \{\hat{\bm{T}}_j\}_{j=1}^{\hat{n}}$ from $\bm{q}$, $\hat{\bm{q}}$ using forward kinematics.\;
		$\bm{D} \gets d(\bm{T}, \hat{\bm{T}}, \alpha_{tr}, \alpha_{rot}) \; \bm{foreach} \; (\bm{T}, \hat{\bm{T}})$ in $\{\bm{T}_i\}_{i=1}^{n} \times \{\hat{\bm{T}}_j\}_{j=1}^{\hat{n}}$\;
		Either calculate correspondence matrix $\bm{W}(\bm{s}, \hat{\bm{s}})$ or use static correspondence matrix $\bm{W}$.\;
		$\bm{D}_W \gets \bm{D} \circ \textbf{W}$\;
		$\overline{\textbf{D}_W} = \frac{\sum_{i=1}^{m}\sum_{j=1}^{n}D_{W,ij}}{m n}$\;
		\textbf{return} $\overline{\textbf{D}_W}$
	}
	\caption{Calculating the Distance Measure}
	\label{alg:distancemeasure}
\end{algorithm}

\section{RESULTS}
\label{chap:Experiments}
We studied static and dynamic imitation tasks in simulation between two dissimilar embodiments using the previously derived distance measure. First, we present results of static pose imitation tasks  between planar manipulators with different links using gradient descent. Second, we examined whether a neural network can learn the optimal static pose between two planar manipulators and, furthermore, between two \textit{Franka Emika Panda} robots for a given expert pose.
In addition, we investigated how well the learner generalized to poses it has not seen during training.
Third, we present results for a dynamic imitation tasks between two \textit{Franka Emika Panda} robotic arms. For this purpose, a simulation environment was built in the physics simulator \textit{Gazebo}. In all simulations, we assumed that the learner has no knowledge of the robot dynamics.

\subsection{Static Pose Imitation Task}
\label{sec:position_imitation}
In a static pose imitation task, the optimal pose of the learner, $\hat{\bm{q}}^\ast $ is obtained for a given pose of the expert, $\bm{q}$, by minimization of the distance measure
\bea
\label{eq:argmin_objective}
\hat{\bm{q}}^\ast = \argmin_{\hat{\bm{q}}} \; d(\bm{s}(\bm{q}), \hat{\bm{s}}(\hat{\bm{q}}))\,.
\eea
Minimizing the distance function is a nonlinear optimization problem for which generally no analytical solution exists, in particular, for embodiments with a large number of links. Using mathematical libraries such as \textit{TensorFlow}, the gradient of the distance function can be computed and local minima can be found numerically via gradient descent.
Instead of trying to solve the optimization problem repeatedly for each input, we can learn a mapping from joint angles of the expert to joint angles of the learner $\bm{f}_{\bm{\theta}}: \bm{q} \longrightarrow \hat{\bm{q}}$, where $\bm{q} \in  [-\pi, \pi]^{n}$ and~$\hat{\bm{q}} \in [-\pi, \pi]^{\hat{n}}$.
The function $\bm{f}_{\bm{\theta}}$ can be approximated by a neural network with weight parameters $\bm{\theta}$. The distance measure was implemented as a computation graph in TensorFlow%
~\cite{abadi2016tensorflow}.

\begin{figure}[]
	\centering
	\begin{subfigure}[b]{0.48\linewidth}
		\includegraphics[width=\textwidth]{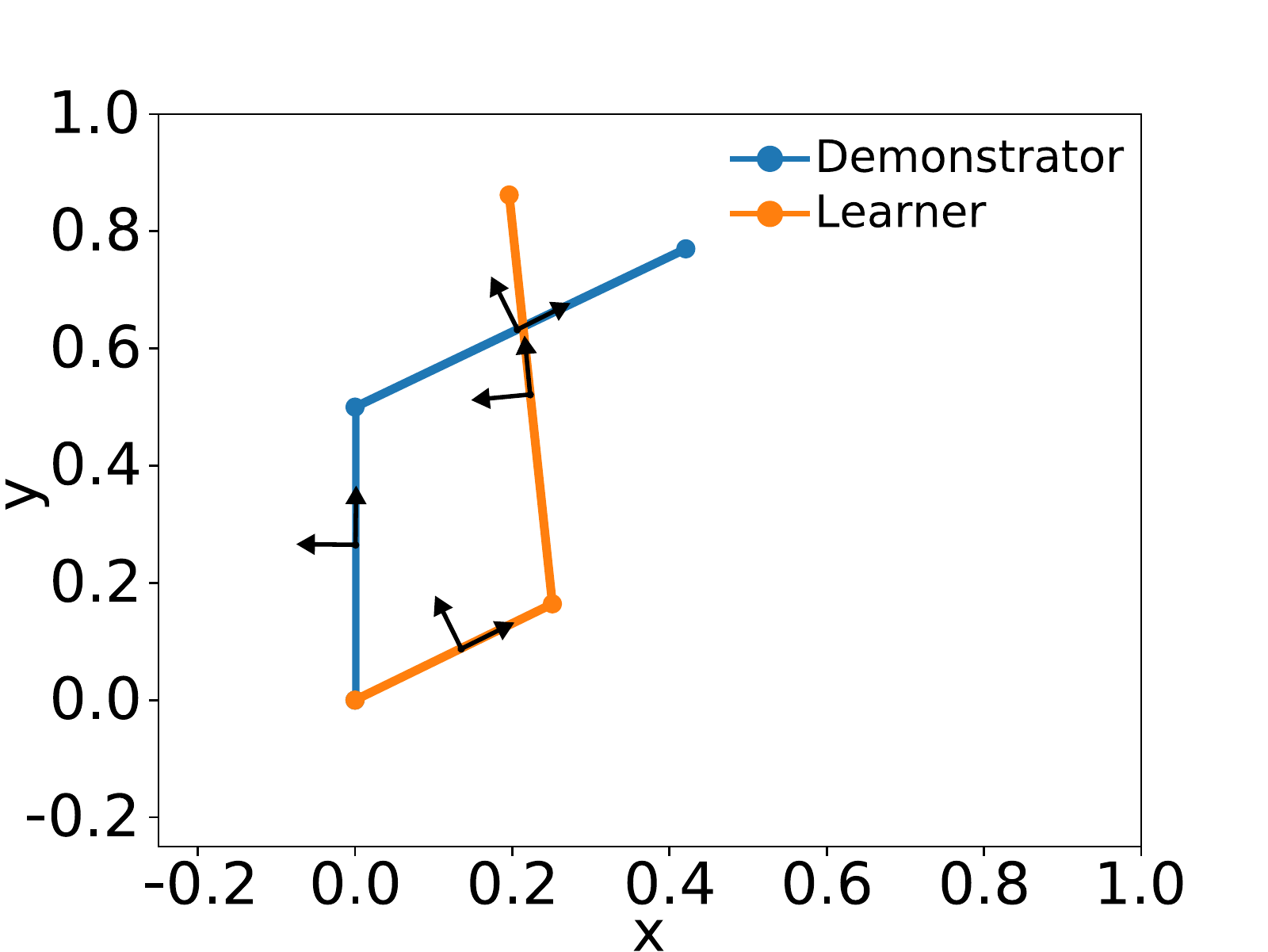}
		\caption{$\alpha_{tr}=0.5,\; \alpha_{rot}=1.0$} \label{fig:influence_weighting:a}
	\end{subfigure}
	\begin{subfigure}[b]{0.48\linewidth}
		\includegraphics[width=\textwidth]{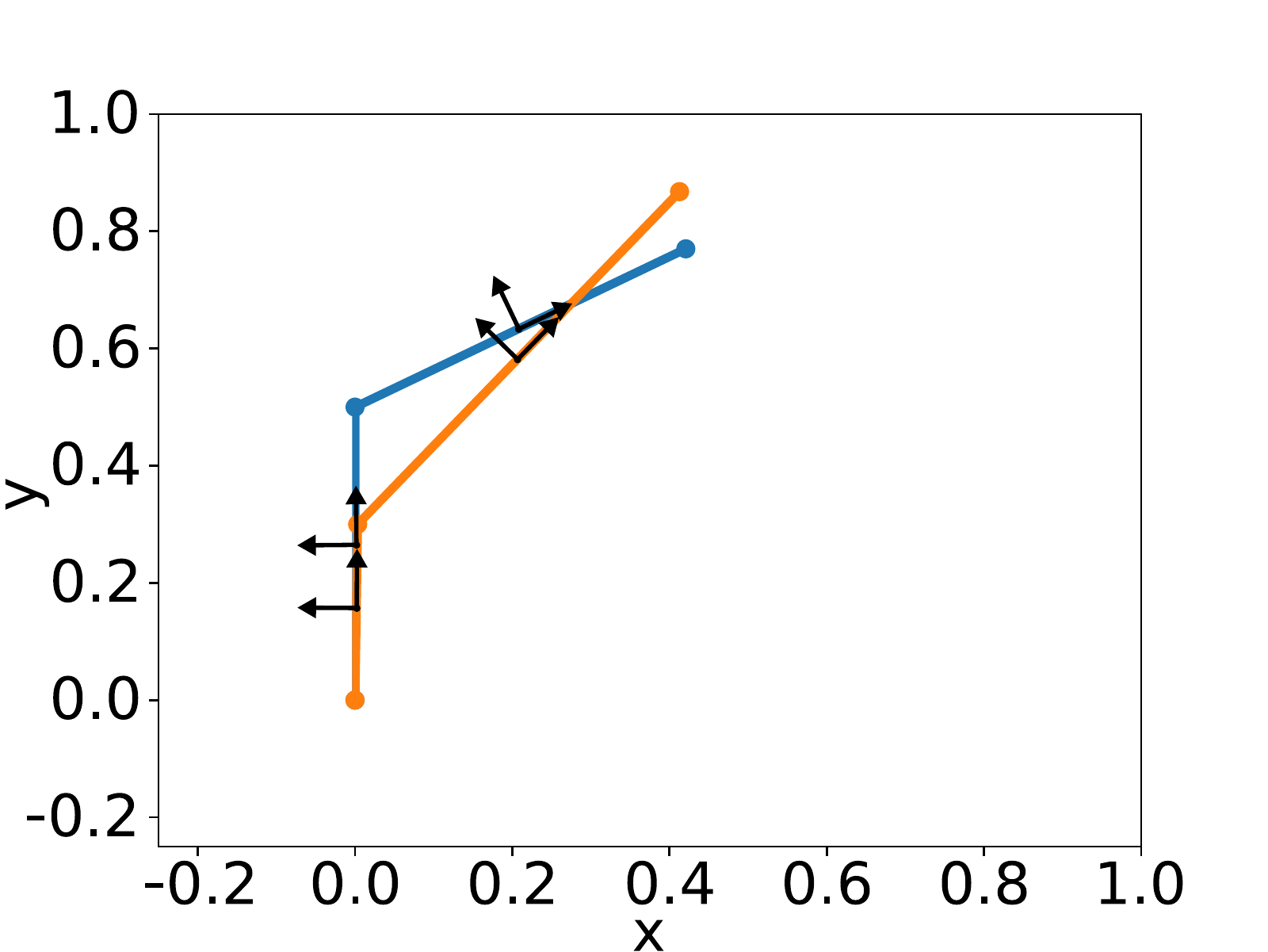}
		\caption{$\alpha_{tr}=3.0,\; \alpha_{rot}=1.0$} \label{fig:influence_weighting:b}
	\end{subfigure}
	\begin{subfigure}[b]{0.48\linewidth}
		\includegraphics[width=\textwidth]{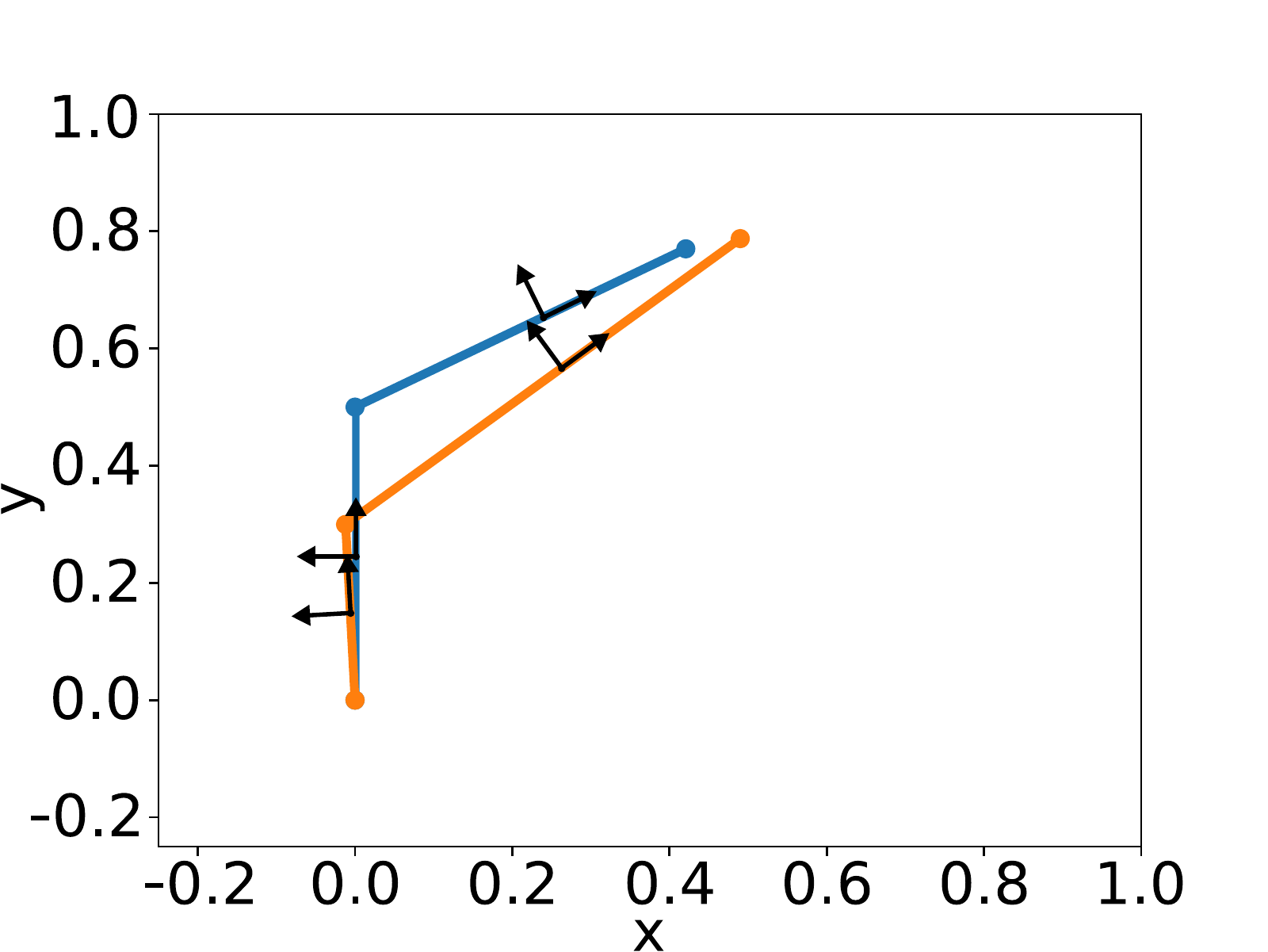}
		\caption{$\alpha_{tr}=1.0,\; \alpha_{rot}=1.0$} \label{fig:influence_weighting:c}
	\end{subfigure}
	\begin{subfigure}[b]{0.48\linewidth}
		\includegraphics[width=\textwidth]{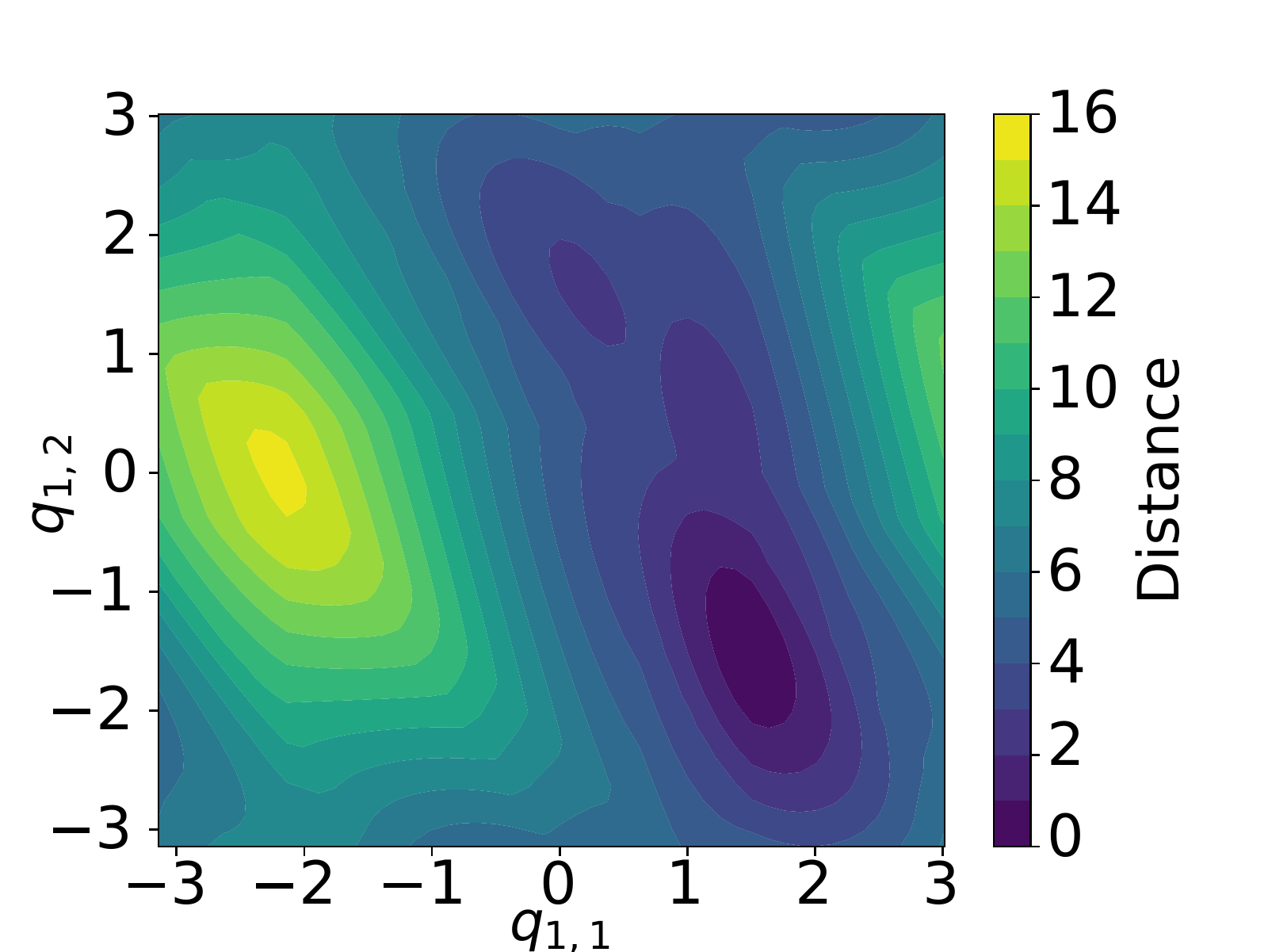}
		\caption{} \label{fig:contours:a}
	\end{subfigure}
	\begin{subfigure}[b]{0.48\linewidth}
		\includegraphics[width=\textwidth]{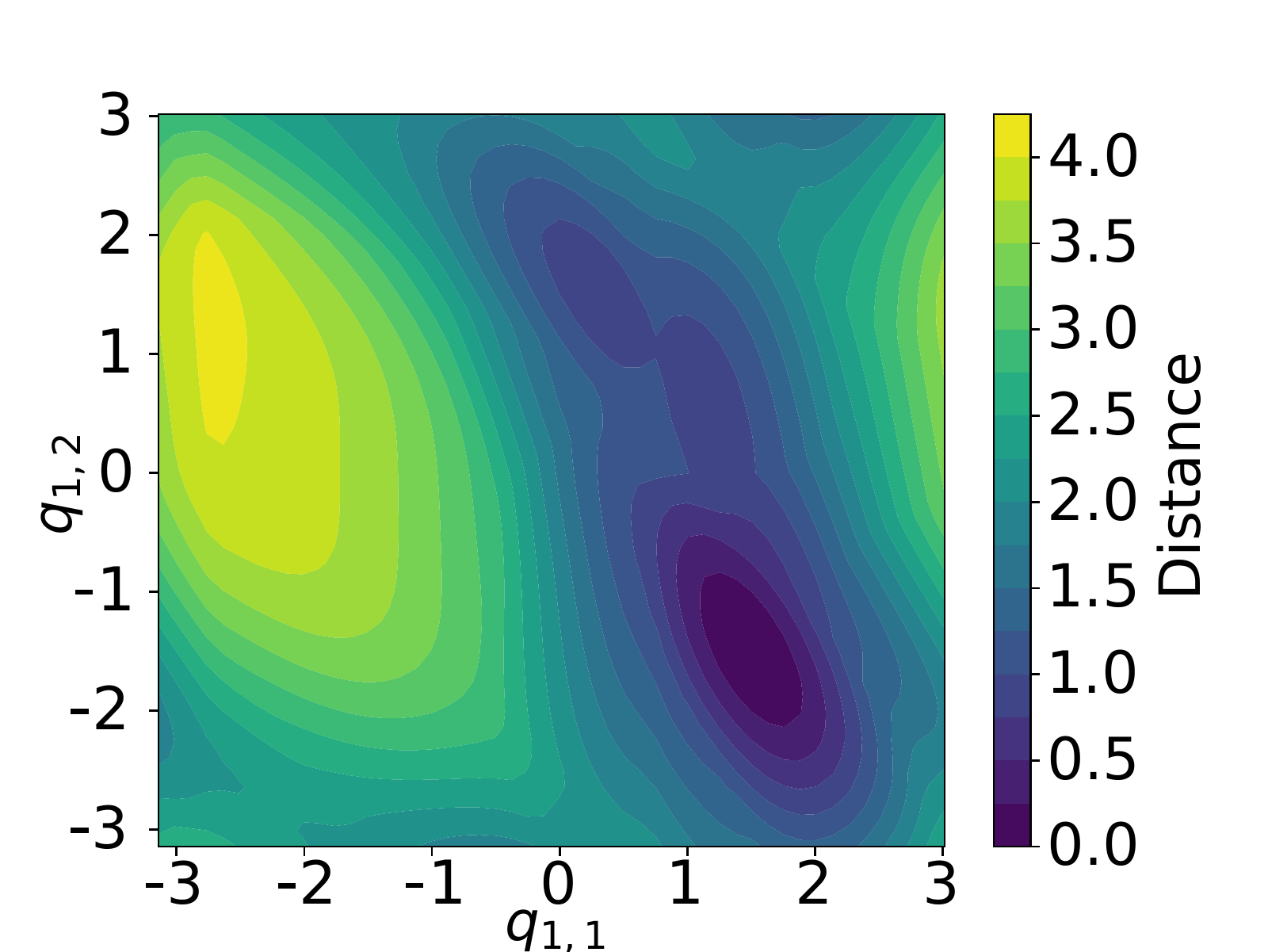}
		\caption{} \label{fig:contours:b}
	\end{subfigure}
	\raggedright
	\begin{subfigure}[b]{0.48\linewidth}
		\includegraphics[width=\textwidth]{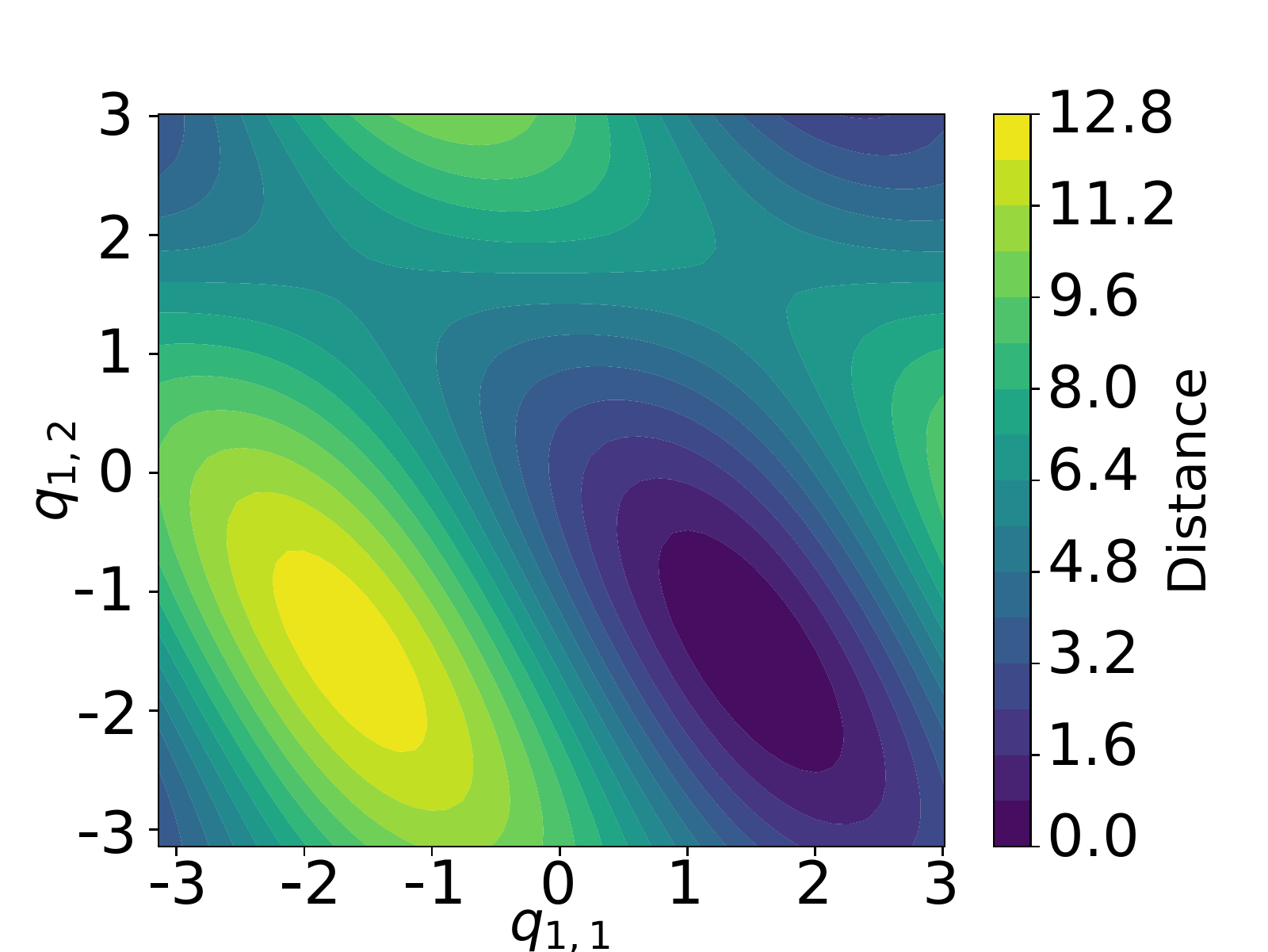}
		\caption{} \label{fig:contours:c}
	\end{subfigure}
	\caption{
		(a-c) Effects of using different distance weighting factors on pose imitation between two planar manipulators (expert: blue, learner: orange).\\	
		\textit{(d-f)} Distance function between planar manipulators with two links. (d) State-dependent weight matrix with $\alpha_{tr}=1.5, \alpha_{rot}=1.0$; (e) State-dependent weight matrix with distance-dependent weighting factors; (f) State-independent weight matrix, considering only rotational distance between corresponding links ($\alpha_{tr}=0$). The distance is plotted over all possible joint angles of one embodiment, while the other manipulator remains fixed at $\bm{q}=[1.5, -1.5]$.}
	\vspace{-0.2cm}
	\label{fig:influence_weighting}
\end{figure}
\textit{Comparing Link Correspondences}. Before training the neural network, we analyzed the behavior of the distance function for a simple toy model.  \autoref{fig:influence_weighting}a-c shows an imitation task between planar manipulators with two links.  The distance was measured using a state-dependent correspondence matrix~$\bm{W}$, using varying weight factors $\alpha_{tr}$,~$\alpha_{rot}$. Each pose of the learner was found via gradient descent. The choice of weight factors clearly has a strong influence on the quality of the result.
Balancing between translational and rotational weights is challenging. One possibility to overcome this difficulty may be to simply use the translational distance as the translational weight and to redefine the rotational weight by subtracting the translational distance from its maximum value and rescaling from $[0, \pi]$ to $[0, 2]$. This way, translational and rotational distance are in the same value range, i.e.,
\begin{equation}
\label{eq:transdependent_alpharot}
\alpha_{tr} = d_{tr}, \quad \alpha_{rot} = \frac{2}{\pi}(2 - d_{tr}).
\end{equation}
The maximum distance results from the fact that both embodiments are normalized to be in a sphere of radius $1$ or diameter $2$, which is the maximum Euclidean distance between two points in this sphere. Equation (\ref{eq:transdependent_alpharot}) ensures that the sum of $\alpha_{tr}$ and~$\alpha_{rot}$ is always $2$, excluding the transformation factor $2/\pi$.
The problem with the above approaches is that there exist local minima in the distance function, as can be observed in \autoref{fig:contours:a} and \ref{fig:contours:b} for two-link manipulators.
\autoref{fig:contours:c} on the other hand, shows that when considering the only the orientational between frames (setting $\alpha_{tr}=0$) using a static, precalculated correspondence matrix, only one minimum remains	. This approach is less flexible but more robust and resulted in parallel alignment of corresponding links. Therefore, all following experiments were conducted using this distance measure. 

\textit{Pose Imitation Mapping Using a Neural Network}. We next implemented a neural network to map joint angles of the expert to corresponding joint angles of the learner, leading to a more efficient method for static pose imitation than conducting a gradient descent search for each state. To generate this nonlinear map, network parameters
\begin{equation}
\label{eq:sgd_objective}
\bm{\theta^*} = \argmin_{\bm{\theta}} \sum_{i=1}^{N}d(\bm{s}(\bm{q}_{i}), \hat{\bm{s}}(f_{\bm{\theta}}(\bm{q}_{i})))\,,
\end{equation}
need to be determined, which minimize the distance between the states of the expert and the states of the learner for a given training set $\{\bm{q}_{1}, \bm{q}_{2}, \dots, \bm{q}_{N}\}$, where the map $\bm{f}_{\bm{\theta}}(\bm{q}_{i})$ is represented by a neural network and~$\bm{\theta}$ are the network parameters. The training dataset $\{\bm{q}_{1}, \bm{q}_{2}, \dots, \bm{q}_{N}\}$ can be generated randomly because it contains only expert angles that do not need to be labeled. The network structure consists of three hidden layers of size $32$ with LReLU activation functions. The output layer uses the \textit{tanh} as activation function, resulting in output values in $[-1, 1]$. These values can then be mapped to angular values in $[-\pi, \pi]$. Having generated a training dataset and another dataset for validation, the network was trained using a minibatch-based stochastic gradient descent method. After dividing the training set into minibatches, the update step is performed for each of the minibatches. Afterwards, the whole training set is shuffled and the procedure repeated.
\begin{figure}[]
	\centering
	\begin{subfigure}[b]{0.48\linewidth}
		\includegraphics[width=\textwidth, clip=true, trim= 0 0 0 20]{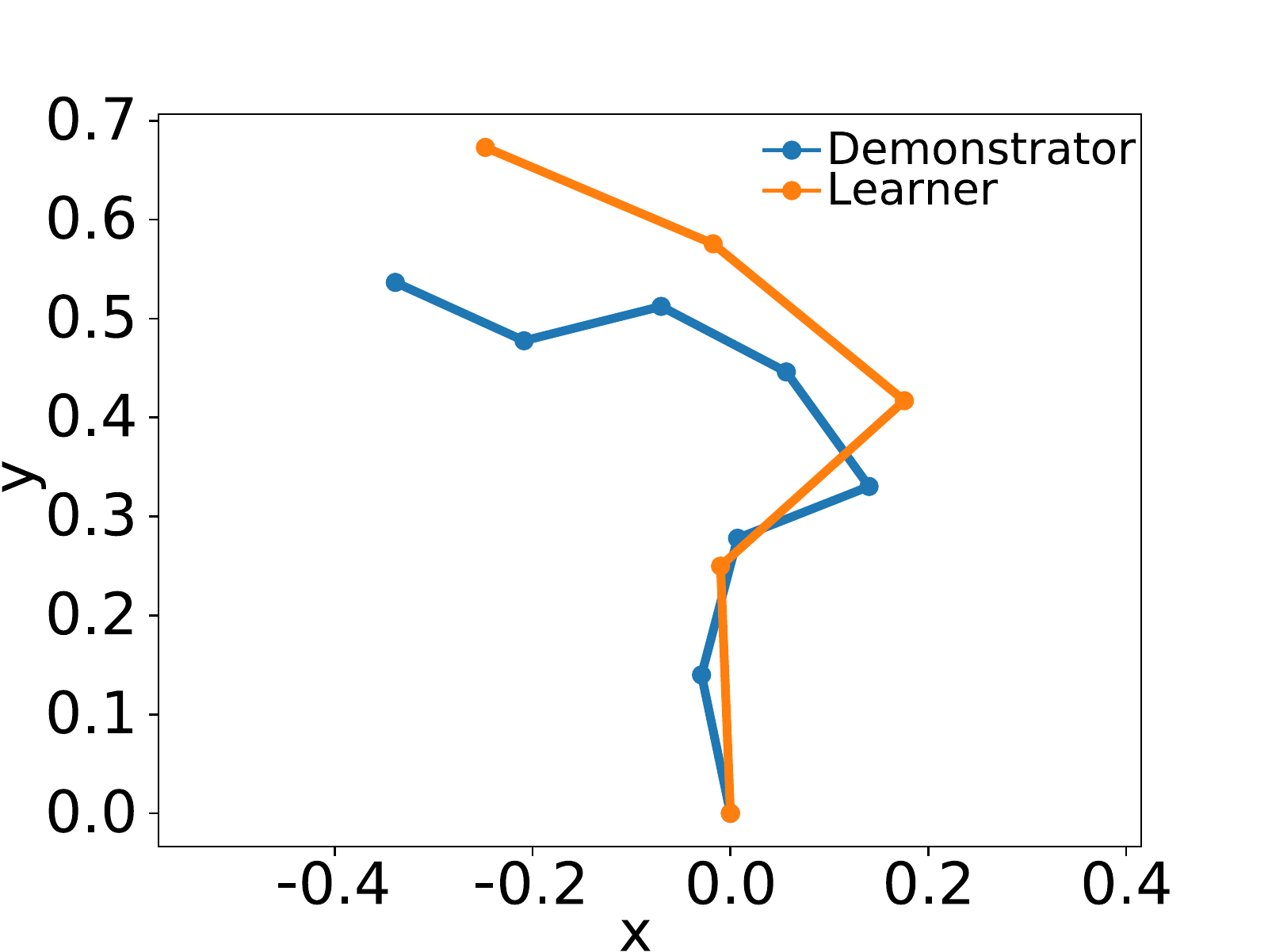}
		\caption{} \label{fig:static_imitation:a}
	\end{subfigure}
	\begin{subfigure}[b]{0.48\linewidth}
		\includegraphics[width=\textwidth, clip=true, trim= 0 0 0 20]{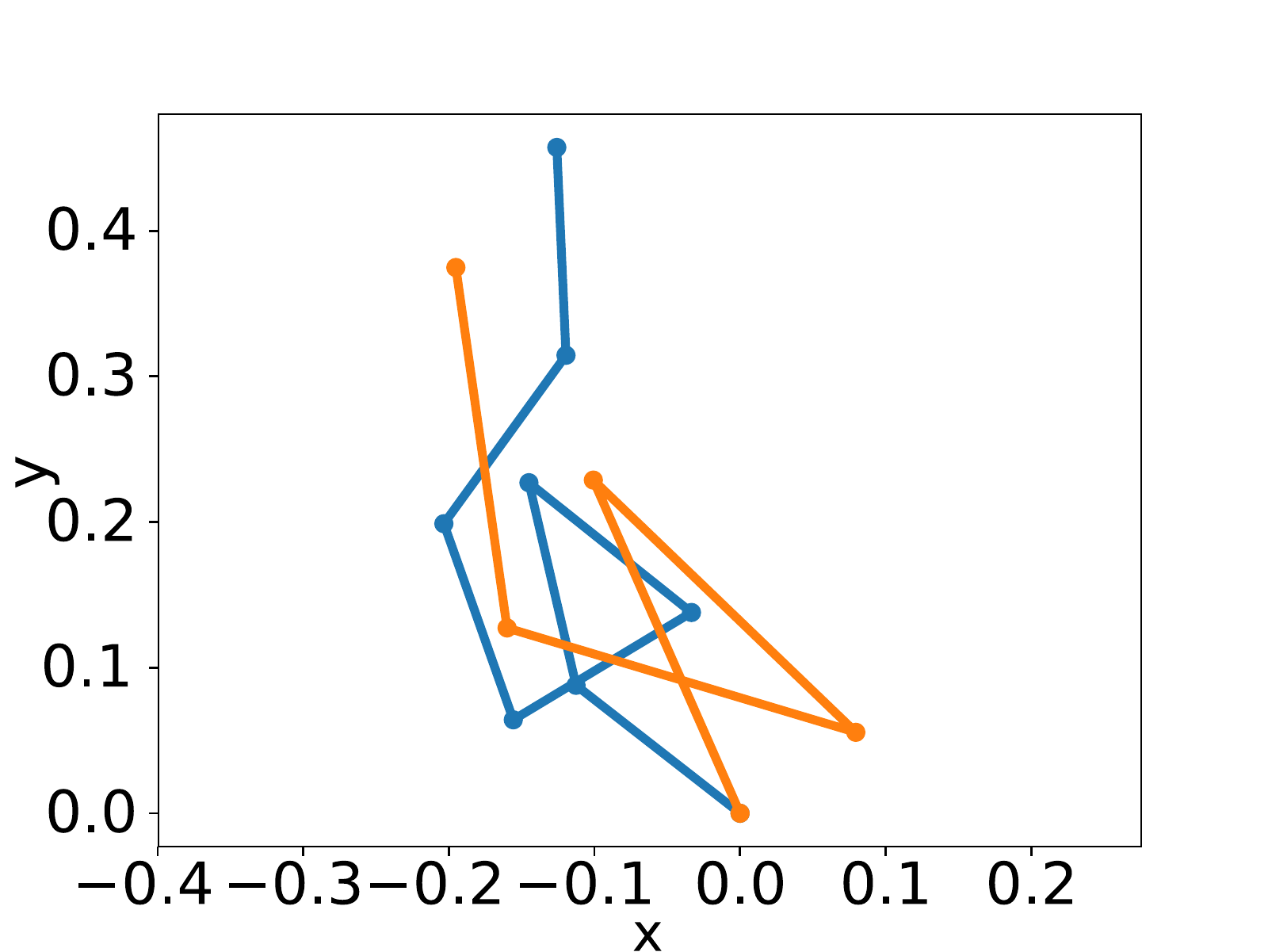}
		\caption{} \label{fig:static_imitation:b}
	\end{subfigure}
	\begin{subfigure}[b]{0.48\linewidth}
		\includegraphics[width=\textwidth, clip=true, trim= 0 0 0 20]{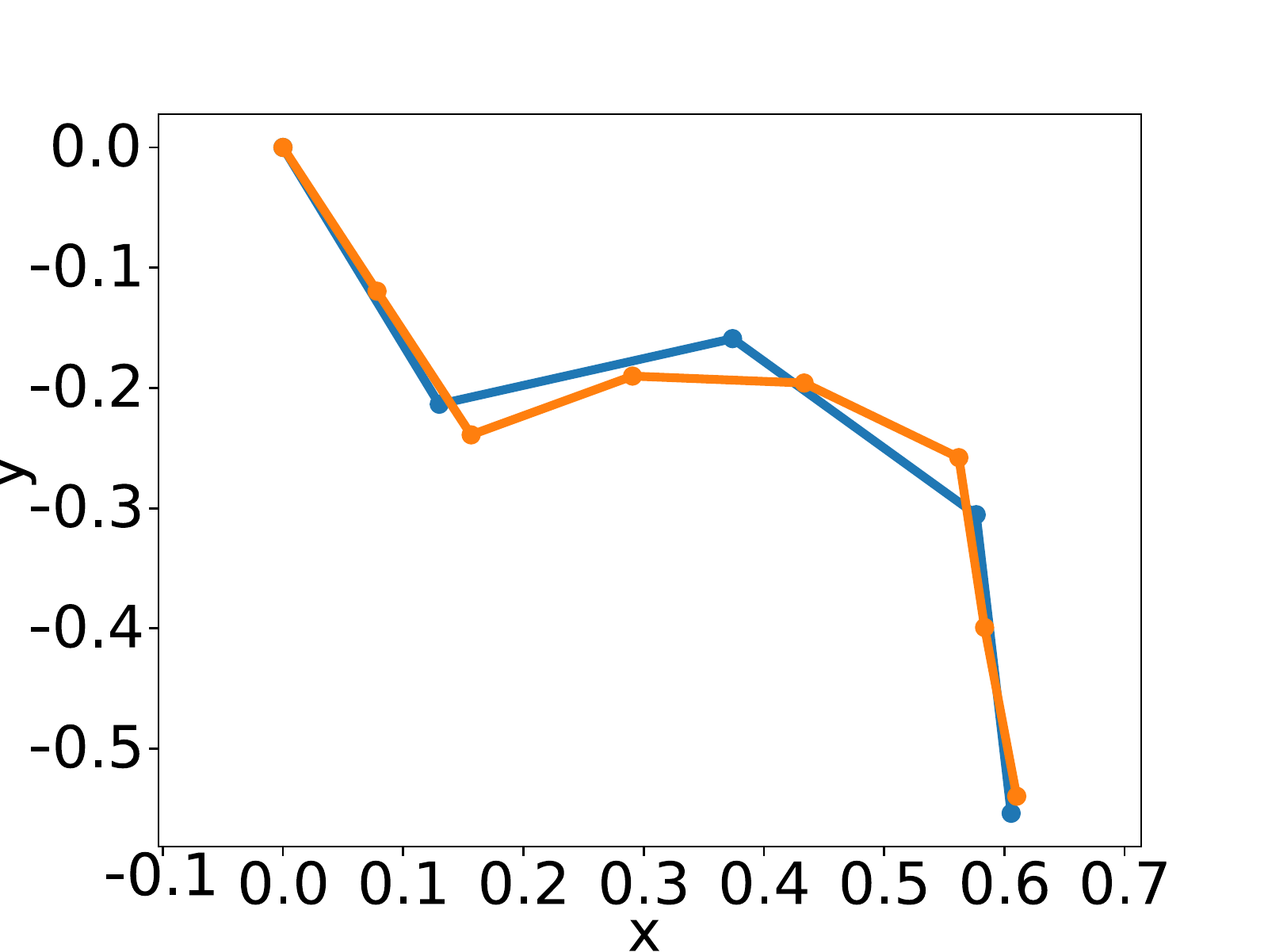}
		\caption{} \label{fig:static_imitation:c}
	\end{subfigure}
	\begin{subfigure}[b]{0.48\linewidth}
		\includegraphics[width=\textwidth, clip=true, trim= 0 0 0 20]{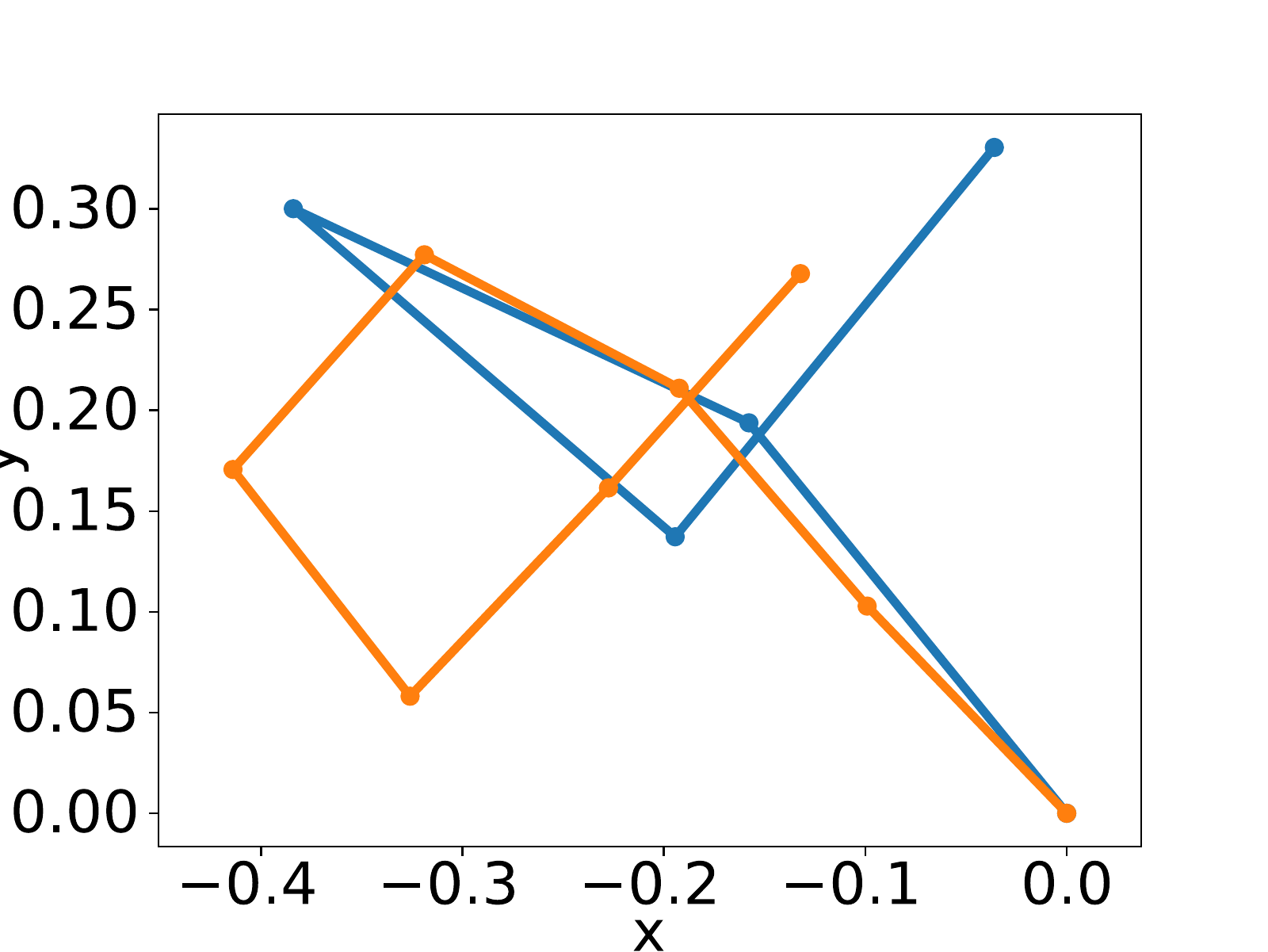}
		\caption{} \label{fig:static_imitation:d}
	\end{subfigure}
	\caption{Imitation between two planar manipulators using a neural network. (a-b) Expert: 7-DOF, learner: 4-DOF; (c-d) Expert: 4-DOF, learner: 7-DOF.}
	\label{fig:static_imitation}
	\vspace{-0.2cm}
\end{figure}
We trained a neural network to find a mapping from joint angles of a 7-DOF expert manipulator to angles of a 4-DOF learner manipulator and another network for the same pair of manipulators but with switched expert/learner roles. Each time we used a training set of 1024 expert demonstrations, dividing it into 32 minibatches in each episode. \autoref{fig:static_imitation} shows the learned poses of the trained network for given expert angles that were \textit{not} included in the training set.

\textit{Pose Imitation Between Three-Dimensional Embodiments}. We next applied the method from the previous section to two Franka Emika Panda robotic arms in simulation. Dissimilar embodiments were generated by locking individual~DOFs of the learner to~$q_i = 0$. For example, to simulate a four-link-Panda robot with four~DOFs, joint 3, 6, and 7 were locked. To simulate a three-link-Panda robot, additionally link 2 was locked (see \autoref{fig:sgd_poses_fourjoint}). We first studied static pose imitation between identical embodiments with all full 7 DOFs enabled (\href{https://youtu.be/UPZclkFoFXQ}{\textit{Video\,1}}\,\footnote{\href{https://youtu.be/UPZclkFoFXQ}{https://youtu.be/UPZclkFoFXQ}}).
\autoref{fig:sgd_poses_fourjoint:a} shows an example of how the trained network solves the task  between dissimilar embodiments (expert: 7 DOFs, learner: 4 DOFs). The locked joints 6 and 7 (indicated in red) lie at the very end of the embodiment and therefore do not contribute much to the overall configuration of the embodiment, in contrast to the locked joint 3 at the center of the embodiment. In \autoref{fig:sgd_poses_fourjoint:b}, the learner only has 3 DOFs. The learner tries to establish similarity by rotating joint 4 as the second joint is locked. The results can be seen in     
 \href{https://youtu.be/BmFH6Nr9F1Y}{\textit{Video\,2}}\,\footnote{\href{https://youtu.be/BmFH6Nr9F1Y}{https://youtu.be/BmFH6Nr9F1Y}}.

The experiments have shown that training a neural network with a distance-based loss function worked reasonably well for static pose imitation.
Local minima in the distance function and over-fitting on the training set posed some problems. While the former  is more difficult to solve, the latter can be solved by increasing the size of the training set and stopping the training process at a suitable time.  Another possibility to improve the network's performance may lie in the structure and training of the network. We used the same structure for all pose imitation tasks without employing techniques that decrease the probability of over-fitting, such as dropout or regularization~\cite{aggarwal2018neural}.

\subsection{Dynamic Motion Imitation}
\label{sec:motion_imitation}
In this section, we a apply a reinforcement learning algorithm for motion imitation by using the online distance measure between embodiments as a feedback signal. Reinforcement learning has shown great success in learning motor tasks, for example, \cite{heess2017locomotion, openai2018dexterity}. We study the transfer of motions from one Panda robot with maximally seven DOFs to another one in simulation. As before, different embodiments are generated by locking DOFs in the learner.
We assume  that the dynamics of the robots are unavailable (model-free) and that the learner is controlled by joint torques $\bm{\tau}$. Consequently, the agent needs to control the joint positions but also, implicitly, learn the the robot dynamics.
\textit{Simulation Environment.} The manufacturer (Franka Emika) of the Panda robot provided a good integration of the robot into the ROS ecosystem, which we augmented by the Gazebo physics simulator.
Unfortunately, no exact inertia values for the Panda were provided, which were needed to simulate the dynamics.
The CoR-Lab group from the  Universität Bielefeld published some estimates of inertia values on their GitHub repository\footnote{https://github.com/corlab/cogimon-gazebo-models/blob/master/franka/robots/panda\_arm.xacro}.
We used these estimates and manually adjusted them by using the  guide given in the Gazebo manual. %
The simulated robot is controlled via joint torque commands. 
To create trajectories of the expert, simple PID-controllers were configured for each joint. Due to the lack of sophisticated controllers and to facilitate the task for the RL agent, gravity was turned off in the simulation.

\textit{RL Environment and Agent.} The next task consisted in the implementation of the reinforcement learning agent 
and the interface for interaction with the simulation. 
The state space consisted of the expert and learner states, thus,
the environment's state $\bm{s}$ is  defined by the tuple $\bm{s}_t = \{\bm{s}(\bm{q}_t, \dot{\bm{q}}_t), \hat{\bm{s}}(\hat{\bm{q}}_t, \dot{\hat{\bm{q}}}_t)\}$. The actions are given by the torque commands of the learner, $\hat{\bm{a}}_t = \hat{\bm{\tau}}_t$, subject to the torque limits for each corresponding joint. The control of the expert is not observable by the RL agent. For training and testing, multiple random trajectories of similar duration were recorded. 
The step size was set to $\Delta t=0.1$s and a step consisted of the following transitions:
The agent  executed an action in the environment by sending a torque command to the learner. The torques were then applied  for the duration of the simulation time $\Delta t$. The simulation then paused and returned the next observed state $s_{t+1}$ together with the  reward $r_{t+1}$, which was calculated from the state as the negative distance measure between the embodiment states.
To train the agent, we used  Proximal Policy Optimization (PPO) \cite{schulman2017proximal},  which is a state-of-the art actor-critic DRL algorithm. One big advantage of PPO is its robustness towards hyperparameter tuning. We employed the GPU-optimized version (PPO2) of the \textit{Stable-Baselines} repository, based on OpenAI's implementations.

\textit{Simulation 1: Imitation of a Single Trajectory}. We first tested, whether motion imitation using reinforcement learning is feasible by transferring a single trajectory between two Panda robots with all 7 DOFs activated, i.e., expert and learner had identical embodiments. Both expert and learner started in their zero pose, in which all joint angles and joint velocities are set to zero. The trajectory of the expert was recorded off-line by moving each joint to an arbitrarily chosen goal position. The environment was reset  whenever the trajectory ended. Each trajectory had a total duration of 5 s, which leads to 50 steps per episode.  
The discount factor was set to $\gamma=0.4$ implying that the agent acted myopically.
This value was chosen because high values led to very slow training progress. The weights for the frame-distance function were set to $\alpha_{tr} = 0.0,\; \alpha_{rot}=1.0,\; \alpha_{v}=0.001,\; \alpha_{\omega}=0.01$ in all simulation experiments.
Training time was about $7.5$ hours on a desktop computer using GPU-accelerated computations.
The simulation showed that the learned trajectory resembles the expert's trajectory very closely and that imitation of motions is possible using a reinforcement learning framework with a distance related reward function. The results can be seen in  \href{https://youtu.be/dLN314VJTHg}{\textit{Video\,3}}\,\footnote{\href{https://youtu.be/dLN314VJTHg}{https://youtu.be/dLN314VJTHg}}.

\textit{Simulation 2: Generalization Between Trajectories.}
\begin{figure}[]
	\centering
	\begin{subfigure}[b]{0.75\linewidth}
		\includegraphics[width=\textwidth, clip=true, trim= 0 20 0 25]{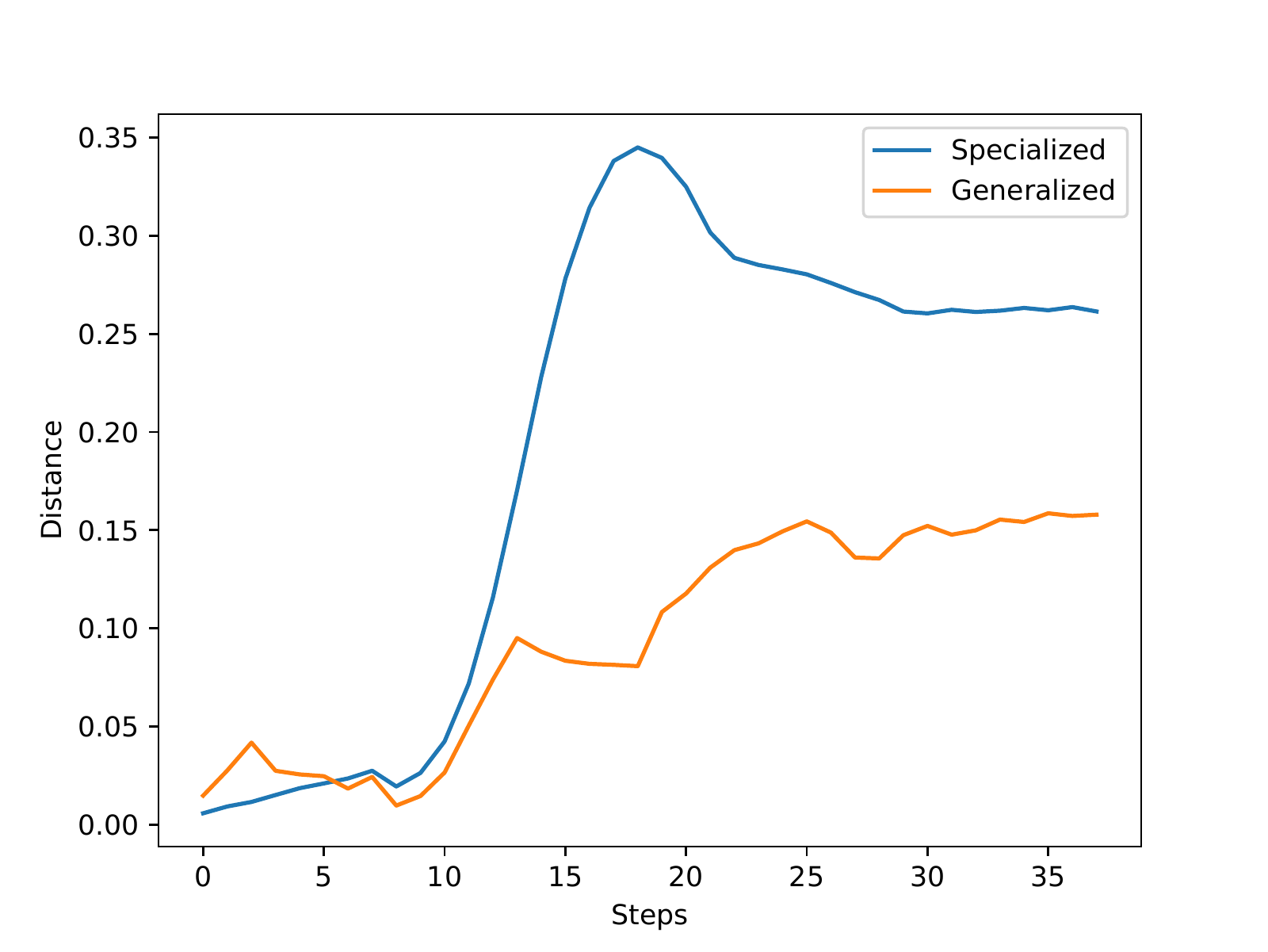}
	\end{subfigure}
	\caption{Generalization capabilities of two different agents. Shown is the distance measure for two 7-DOF-Panda learners while imitating the same, previously unseen trajectory. One learner (''Specialized``) was trained on a single trajectory, whereas the other (''Generalized``) was trained on~124 different trajectories.}
	\label{fig:rl_e7_l7_performance_unseen}
	\vspace{-0.2cm}
\end{figure}
It was examined next, whether the agent can imitate trajectories it had not seen before. For this purpose, we trained the agent with different numbers of training data (one  vs. 124 trajectories). Trajectories were recorded as before, but each time with different final poses. 
The trajectory of the agent trained on a single trajectory barely resembles the trajectory of the expert. The imitation of the agent trained on the larger data set is not perfect, but  resembles the trajectories from the expert more closely (see \href{https://youtu.be/2Q7jiY9DRUg}{\textit{Video\,4}}\,\footnote{\href{https://youtu.be/2Q7jiY9DRUg}{https://youtu.be/2Q7jiY9DRUg}}). This improvement shows in a significantly smaller distance between expert and learner along the trajectory (see \autoref{fig:rl_e7_l7_performance_unseen}).

\textit{Simulation 3: Imitation Between Dissimilar Embodiments.}
In the next experiment, we studied  motion transfer between dissimilar Panda robots. Towards this goal, we trained the agent again on the same 124 trajectories but this time the learner had only 4 or 3 DOFs, respectively.
As the learning robot was restricted in its DOFs, some trajectories could could not be imitated well, resulting in higher values of the distance measure.
We found that the restricted learner moved its links in similar directions as the expert but the restrictions  prevented a more similar imitation. Examples can be seen in \href{https://youtu.be/Fytw8sz0pG0}{\textit{Video\,5}}\,\footnote{\href{https://youtu.be/Fytw8sz0pG0}{https://youtu.be/Fytw8sz0pG0}}.

\section{CONCLUSIONS}
Our main contributions with this work are threefold:
First, we have introduced a definition of embodiment states in terms of frames/twists and candidate points. Second, we have povided a distance measure between dissimilar embodiments using correspondences between frames  of expert and learner. Third, we have applied this distance measure to static pose and movement imitation tasks between manipulators. All tasks have been performed in simulation. In all experiments we could show that the agent was able to learn the imitation task, even though no dynamic model has been provided to the learner. The framework that we have developed is generic and flexible and not limited to our choice of parameters, distance measures and type of robots. Depending on the correspondence matrix calculation, the topology of the embodiments is not crucial. Possibly even free topologies like swarms of flying objects could be compared and brought into similarity.

\vspace*{-0.5cm}
\bibliographystyle{IEEEtran}
\bibliography{references}

\begin{thebibliography}{10}
\providecommand{\url}[1]{#1}
\csname url@rmstyle\endcsname
\providecommand{\newblock}{\relax}
\providecommand{\bibinfo}[2]{#2}
\providecommand\BIBentrySTDinterwordspacing{\spaceskip=0pt\relax}
\providecommand\BIBentryALTinterwordstretchfactor{4}
\providecommand\BIBentryALTinterwordspacing{\spaceskip=\fontdimen2\font plus
\BIBentryALTinterwordstretchfactor\fontdimen3\font minus
  \fontdimen4\font\relax}
\providecommand\BIBforeignlanguage[2]{{%
\expandafter\ifx\csname l@#1\endcsname\relax
\typeout{** WARNING: IEEEtran.bst: No hyphenation pattern has been}%
\typeout{** loaded for the language `#1'. Using the pattern for}%
\typeout{** the default language instead.}%
\else
\language=\csname l@#1\endcsname
\fi
#2}}

\bibitem{abbeel2010autonomous}
P.~Abbeel, A.~Coates, and A.~Y. Ng, ``Autonomous helicopter aerobatics through
  apprenticeship learning,'' \emph{The International Journal of Robotics
  Research}, vol.~29, no.~13, pp. 1608--1639, 2010.

\bibitem{kober2010imitation}
J.~Kober and J.~Peters, ``Imitation and reinforcement learning,'' \emph{IEEE
  Robotics \& Automation Magazine}, vol.~17, no.~2, pp. 55--62, 2010.

\bibitem{kormushev2011imitation}
P.~Kormushev, S.~Calinon, and D.~G. Caldwell, ``Imitation learning of
  positional and force skills demonstrated via kinesthetic teaching and haptic
  input,'' \emph{Advanced Robotics}, vol.~25, no.~5, pp. 581--603, 2011.

\bibitem{boularias2011relative}
A.~Boularias, J.~Kober, and J.~Peters, ``Relative entropy inverse reinforcement
  learning,'' in \emph{Proceedings of the Fourteenth International Conference
  on Artificial Intelligence and Statistics}, 2011, pp. 182--189.

\bibitem{calinon2010learning}
S.~Calinon, F.~D'halluin, E.~L. Sauser, D.~G. Caldwell, and A.~G. Billard,
  ``Learning and reproduction of gestures by imitation,'' \emph{IEEE Robotics
  \& Automation Magazine}, vol.~17, no.~2, pp. 44--54, 2010.

\bibitem{asfour2008imitation}
T.~Asfour, P.~Azad, F.~Gyarfas, and R.~Dillmann, ``Imitation learning of
  dual-arm manipulation tasks in humanoid robots,'' \emph{International Journal
  of Humanoid Robotics}, vol.~5, no.~02, pp. 183--202, 2008.

\bibitem{lopes2007affordance}
M.~Lopes, F.~S. Melo, and L.~Montesano, ``Affordance-based imitation learning
  in robots,'' in \emph{2007 IEEE/RSJ International Conference on Intelligent
  Robots and Systems}.\hskip 1em plus 0.5em minus 0.4em\relax IEEE, 2007, pp.
  1015--1021.

\bibitem{ratliff2007imitation}
N.~Ratliff, J.~A. Bagnell, and S.~S. Srinivasa, ``Imitation learning for
  locomotion and manipulation,'' in \emph{2007 7th IEEE-RAS International
  Conference on Humanoid Robots}.\hskip 1em plus 0.5em minus 0.4em\relax IEEE,
  2007, pp. 392--397.

\bibitem{chalodhorn2010learning}
R.~Chalodhorn, D.~B. Grimes, K.~Grochow, and R.~P. Rao, ``Learning to walk by
  imitation in low-dimensional subspaces,'' \emph{Advanced Robotics}, vol.~24,
  no. 1-2, pp. 207--232, 2010.

\bibitem{osa2018algorithmic}
T.~Osa, J.~Pajarinen, G.~Neumann, J.~A. Bagnell, P.~Abbeel, J.~Peters,
  \emph{et~al.}, ``An algorithmic perspective on imitation learning,''
  \emph{Foundations and Trends{\textregistered} in Robotics}, vol.~7, no. 1-2,
  pp. 1--179, 2018.

\bibitem{billard2016learning}
A.~Billard, S.~Calinon, and R.~Dillmann, ``Learning from humans,''
  \emph{Springer Handbook of Robotics}, pp. 1995--2014, 01 2016.

\bibitem{englert2013probabilistic}
P.~Englert, A.~Paraschos, M.~P. Deisenroth, and J.~Peters, ``Probabilistic
  model-based imitation learning,'' \emph{Adaptive Behavior}, vol.~21, no.~5,
  pp. 388--403, 2013.

\bibitem{ross2011reduction}
S.~Ross, G.~Gordon, and D.~Bagnell, ``A reduction of imitation learning and
  structured prediction to no-regret online learning,'' in \emph{Proceedings of
  the fourteenth international conference on artificial intelligence and
  statistics}, 2011, pp. 627--635.

\bibitem{ross2013learning}
S.~Ross, N.~Melik-Barkhudarov, K.~S. Shankar, A.~Wendel, D.~Dey, J.~A. Bagnell,
  and M.~Hebert, ``Learning monocular reactive uav control in cluttered natural
  environments,'' \emph{2013 IEEE International Conference on Robotics and
  Automation}, 2013.

\bibitem{ijspeert2002movement}
A.~J. Ijspeert, J.~Nakanishi, and S.~Schaal, ``Movement imitation with
  nonlinear dynamical systems in humanoid robots,'' in \emph{Proceedings 2002
  IEEE International Conference on Robotics and Automation (Cat. No.
  02CH37292)}, vol.~2.\hskip 1em plus 0.5em minus 0.4em\relax IEEE, 2002, pp.
  1398--1403.

\bibitem{maeda2017phase}
G.~Maeda, M.~Ewerton, G.~Neumann, R.~Lioutikov, and J.~Peters, ``Phase
  estimation for fast action recognition and trajectory generation in
  human--robot collaboration,'' \emph{The International Journal of Robotics
  Research}, vol.~36, no. 13-14, pp. 1579--1594, 2017.

\bibitem{osa2014trajectory}
T.~Osa, K.~Harada, N.~Sugita, and M.~Mitsuishi, ``Trajectory planning under
  different initial conditions for surgical task automation by learning from
  demonstration,'' in \emph{2014 IEEE International Conference on Robotics and
  Automation (ICRA)}.\hskip 1em plus 0.5em minus 0.4em\relax IEEE, 2014, pp.
  6507--6513.

\bibitem{peng2018deepmimic}
X.~B. Peng, P.~Abbeel, S.~Levine, and M.~van~de Panne, ``Deepmimic:
  Example-guided deep reinforcement learning of physics-based character
  skills,'' \emph{ACM Transactions on Graphics (TOG)}, vol.~37, no.~4, pp.
  1--14, 2018.

\bibitem{abbeel2004apprenticeship}
P.~Abbeel and A.~Y. Ng, ``Apprenticeship learning via inverse reinforcement
  learning,'' in \emph{Proceedings of the twenty-first international conference
  on Machine learning}, 2004.

\bibitem{ratliff2006margin}
N.~D. Ratliff, J.~A. Bagnell, and M.~A. Zinkevich, ``Maximum margin planning,''
  in \emph{Proceedings of the 23rd international conference on Machine
  learning}, 2006, pp. 729--736.

\bibitem{silver2010navigation}
D.~Silver, J.~Bagnell, and A.~Stentz, ``Learning from demonstration for
  autonomous navigation in complex unstructured terrain,'' \emph{I. J. Robotic
  Res.}, vol.~29, pp. 1565--1592, 10 2010.

\bibitem{ziebart2008entropy}
B.~Ziebart, A.~Maas, J.~Bagnell, and A.~Dey, ``Maximum entropy inverse
  reinforcement learning.''\hskip 1em plus 0.5em minus 0.4em\relax AAAI Press,
  2008, pp. 1433--1438.

\bibitem{finn2016guided}
C.~Finn, S.~Levine, and P.~Abbeel, ``Guided cost learning: Deep inverse optimal
  control via policy optimization,'' \emph{Proceedings of the 33Rd
  International Conference on Machine Learning}, 03 2016.

\bibitem{ho2016model}
J.~Ho, J.~Gupta, and S.~Ermon, ``Model-free imitation learning with policy
  optimization,'' in \emph{International Conference on Machine Learning}, 2016,
  pp. 2760--2769.

\bibitem{nehaniv2001likeme}
C.~L. Nehaniv and K.~Dautenhahn, ``Like me?- {M}easures of correspondence and
  imitation,'' \emph{Cybernetics and Systems}, vol.~32, pp. 11--51, 2001.

\bibitem{alissandrakis2002alice}
A.~Alissandrakis, C.~Nehaniv, and K.~Dautenhahn, ``Imitating with alice:
  Learning to imitate corresponding actions across dissimilar embodiments,''
  \emph{IEEE Transactions on Systems, Man, \& Cybernetics, Part A: Systems and
  Humans}, vol.~32, pp. 482--496, 07 2002.

\bibitem{alissandrakis2002do}
A.~Alissandrakis, C.~L. Nehaniv, and K.~Dautenhahn, ``Do as {I} do:
  Correspondences across different robotic embodiments,'' in \emph{Procs. 5th
  German Workshop on Artificial Life. Lubeck}, 2002.

\bibitem{alissandrakis2005corresponding}
A.~Alissandrakis, C.~L. Nehaniv, K.~Dautenhahn, and J.~Saunders, ``Achieving
  corresponding effects on multiple robotic platforms: Imitating in context
  using different effect metrics,'' in \emph{In: Proceedings of the Third
  International Symposium on Imitation in Animals and Artifacts}.\hskip 1em
  plus 0.5em minus 0.4em\relax AISB, 2005.

\bibitem{alissandrakis2007correspondence}
A.~Alissandrakis, C.~L. Nehaniv, and K.~Dautenhahn, ``Correspondence mapping
  induced state and action metrics for robotic imitation,'' \emph{IEEE
  Transactions on Systems, Man, and Cybernetics, Part B (Cybernetics)},
  vol.~37, pp. 299--307, 2007.

\bibitem{nehaniv2007book}
C.~L. Nehaniv and K.~E. Dautenhahn, \emph{Imitation and social learning in
  robots, humans and animals: behavioural, social and communicative
  dimensions.}\hskip 1em plus 0.5em minus 0.4em\relax Cambridge University
  Press, 2007.

\bibitem{grimes2006dynamic}
D.~B. Grimes, R.~Chalodhorn, and R.~P. Rao, ``Dynamic imitation in a humanoid
  robot through nonparametric probabilistic inference.'' in \emph{Robotics:
  science and systems}.\hskip 1em plus 0.5em minus 0.4em\relax Citeseer, 2006,
  pp. 199--206.

\bibitem{grimes2009learning}
D.~B. Grimes and R.~P. Rao, ``Learning actions through imitation and
  exploration: Towards humanoid robots that learn from humans,'' in
  \emph{Creating Brain-Like Intelligence}.\hskip 1em plus 0.5em minus
  0.4em\relax Springer, 2009, pp. 103--138.

\bibitem{parkLie}
F.~Park, J.~Bobrow, and S.~Ploen, ``A lie group formulation of robot
  dynamics,'' \emph{The International Journal of Robotics Research}, vol.~14,
  no.~6, pp. 609--618, 1995.

\bibitem{lynch2017modern}
K.~M. Lynch and F.~C. Park, \emph{Modern Robotics}.\hskip 1em plus 0.5em minus
  0.4em\relax Cambridge University Press, 2017.

\bibitem{abadi2016tensorflow}
M.~Abadi, P.~Barham, J.~Chen, Z.~Chen, A.~Davis, J.~Dean, M.~Devin,
  S.~Ghemawat, G.~Irving, M.~Isard, M.~Kudlur, J.~Levenberg, R.~Monga,
  S.~Moore, D.~G. Murray, B.~Steiner, P.~Tucker, V.~Vasudevan, P.~Warden,
  M.~Wicke, Y.~Yu, and X.~Zheng, ``Tensorflow: a system for large-scale machine
  learning,'' in \emph{12th {USENIX} Symposium on Operating Systems Design and
  Implementation ({OSDI} 16)}.\hskip 1em plus 0.5em minus 0.4em\relax {USENIX}
  Association, 2016, pp. 265--283.

\bibitem{aggarwal2018neural}
C.~C. Aggarwal, \emph{Neural networks and deep learning}.\hskip 1em plus 0.5em
  minus 0.4em\relax Springer, 2018, vol.~10.

\bibitem{heess2017locomotion}
N.~Heess, D.~TB, S.~Sriram, J.~Lemmon, J.~Merel, G.~Wayne, Y.~Tassa, T.~Erez,
  Z.~Wang, S.~Eslami, \emph{et~al.}, ``Emergence of locomotion behaviours in
  rich environments,'' \emph{arXiv preprint arXiv:1707.02286}, 2017.

\bibitem{openai2018dexterity}
O.~M. Andrychowicz, B.~Baker, M.~Chociej, R.~Jozefowicz, B.~McGrew,
  J.~Pachocki, A.~Petron, M.~Plappert, G.~Powell, A.~Ray, \emph{et~al.},
  ``Learning dexterous in-hand manipulation,'' \emph{The International Journal
  of Robotics Research}, vol.~39, no.~1, pp. 3--20, 2020.

\bibitem{schulman2017proximal}
J.~Schulman, F.~Wolski, P.~Dhariwal, A.~Radford, and O.~Klimov, ``Proximal
  policy optimization algorithms,'' \emph{arXiv preprint arXiv:1707.06347},
  2017.

\end{thebibliography}

\end{document}